\documentclass[12pt]{article}
\newcommand{\D}{\mathcal{D}}

\def\eqref#1{equation~\ref{#1}}

\def\1{\bm{1}}

\DeclareMathAlphabet{\mathsfit}{\encodingdefault}{\sfdefault}{m}{sl}
\SetMathAlphabet{\mathsfit}{bold}{\encodingdefault}{\sfdefault}{bx}{n}

\usepackage{titling}
\usepackage{url}
\usepackage{hyperref}
\usepackage{natbib}
\hypersetup{
    colorlinks,
    linkcolor={red!50!black},
    citecolor={blue!50!black},
    urlcolor={blue!80!black}
}
\usepackage[nottoc, notlot, notlof]{tocbibind}
\usepackage[caption=false,font=footnotesize]{subfig}
\usepackage{amssymb,amsmath}
\usepackage{pdflscape}
\usepackage{prettyref}
\usepackage{verbatim}
\usepackage[section]{placeins}
\usepackage{csquotes,graphicx}
\usepackage{subcaption} 
\usepackage{multirow} 
\usepackage[table,xcdraw]{xcolor}
\usepackage{longtable}
\emergencystretch=\maxdimen
\usepackage[colorinlistoftodos]{todonotes}

\usepackage{longtable,booktabs}
\IfFileExists{footnote.sty}{\usepackage{footnote}\makesavenoteenv{long table}}{}
\usepackage{graphicx,grffile,booktabs}
\usepackage[parfill]{parskip}

\title{Data Study Group\\Final Report}

\author{
  Mastercard
}

\date{20-24 May 2024}

\makeatletter
\usepackage{Turing-report}

\usepackage{helvet}

\begin{document}

\begin{titlepage}

\noindent
\begin{tikzpicture}[overlay]
	\useasboundingbox (0,0) rectangle(\paperwidth,\paperheight);
	\fill[color=cornerleaf] (0,-10) rectangle (\paperwidth,\paperheight);
	\draw[fill=bkg,color=bkg] (0,-10) -- (0 ,8) -- (1,8) -- (13,3) -- (20,-8) -- (20,-10) -- cycle;	
	\ifnum\thepage>1\relax%
		\fill[bkg,opacity=1] (0,0) rectangle(\paperwidth,\paperheight);
	 \fi
\end{tikzpicture}
\includegraphics[width=0.45\textwidth]{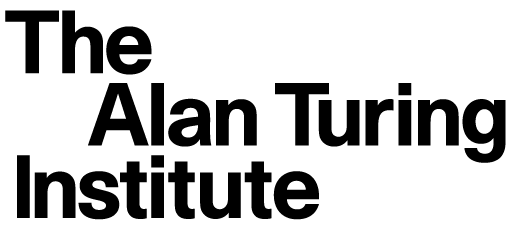}

\vspace{3cm}
\begin{center}
\begin{huge}
{\bf \thetitle }\\
\end{huge}
\vspace{1.5cm}
\begin{Large}
	{\bf \thedate }
\end{Large}
\vspace{1.5cm} \\
\begin{huge}
{\bf \theauthor} \\
\end{huge}
\vspace{0.5cm} 

\begin{LARGE}
{Measuring Fairness in Financial Transaction Machine Learning Models} \\
\end{LARGE}
\end{center}
\end{titlepage}

\setcounter{tocdepth}{3}
\tableofcontents

\newpage
\section{Executive Summary}\label{executive-summary}

\subsection{Challenge and Objectives}
Mastercard, a global leader in financial services, develops and deploys machine learning models aimed at optimizing card usage and preventing attrition through advanced predictive models. These models use aggregated and anonymized card usage patterns, including cross-border transactions and industry-specific spending, to tailor bank offerings and maximize revenue opportunities. Mastercard has established an AI Governance program, based on its Data and Tech Responsibility Principles, to evaluate any built and bought AI for efficacy, fairness, and transparency. As part of this effort, Mastercard has sought expertise from the Turing Institute through a Data Study Group to better assess fairness in more complex AI/ML models. The Data Study Group challenge lies in defining, measuring, and mitigating fairness in these predictions, which can be complex due to the various interpretations of fairness, gaps in the research literature, and ML-operations challenges.

\subsection{Overview of the Data}

Mastercard provided participants with access to synthetic datasets based on financial transactions by customers in a 12-month period~\footnote{The synthetic data does not represent the spending of actual Mastercard cardholders, and subsequent references to "customer" in this paper refers to the synthetic customers created specifically for this project.}. The dataset consists of 1 million rows, 20 features, and 9 output labels.  Each row contains statistics on the overall spending of every customer, along with labels corresponding to different industries~\footnote{The models developed by the DSG participants and presented in this report differ from those developed by Mastercard. They use a smaller subset of variables and a different model architecture with synthetic data. Therefore, the results do not reflect the predictions or data that Mastercard shares with its customers.}.

We have two datasets:

\begin{itemize}
    \item \textbf{Adoption} describes the binary value of whether the customer will spend within a particular industry in the next three months for the first time.

    \item \textbf{Spending} consists of labels specifying the amount of spending by the customer in a particular industry category in the next three months.
\end{itemize}

Finally, for every customer, we have additional information on demographics described by three attributes  that can be a source of potential multidimensional discrimination ~\citep{DBLP:conf/fat/0001HN23}\footnote{These features are not used in the original model but will be employed during this project to assess  bias}:
\begin{itemize}
    \item \textbf{Gender:} Female, male, or unspecified
    \item \textbf{Ethnicity:} White, Hispanic, Black, Asian, Middle Eastern, Native Indian, or not given/unspecified.
    \item \textbf{Age:} Below or above 40 years old
\end{itemize}

The model predictions span over multiple industry labels (multi-label), and we assess fairness for different dimensions (gender, ethnicity and age) (intersectional fairness). This leads to a complex intersectional multi-label fairness challenge.

\subsection{Structure and Content}

Mastercard's current approach uses multi-label classification and regression techniques to predict spending behaviors.
The fairness methodology for this project is structured into several distinct phases, each addressing different aspects of the project:

\textbf{Understanding Fairness in Financial Transactions} (Section~\ref{sec:background}): This initial phase involves discussions with the Mastercard team to define the business problem, identify fairness requirements, and review previous related bias assessment efforts.

\textbf{Data Overview} (Section~\ref{sec:data}): We provide a detailed description of the data set used, along with an exploratory data analysis to understand the underlying patterns and characteristics of the data.

\textbf{Mathematical Formulation} (Section~\ref{sec:math.formulation}): We develop a mathematical framework suited for addressing intersectional fairness in multi-label problems, tailored specifically to this scenario.

\textbf{Measuring Discrimination} (Section~\ref{sec:measuring}): This phase involves assessing fairness within the application, using defined metrics to evaluate how the models perform across different demographic groups.

\textbf{Mitigating Fairness} (Section~\ref{sec:mitigation}): We implement strategies to mitigate bias, employing both in-processing and post-processing techniques to improve fairness in the machine learning models.

\textbf{Limitations and Opportunities} (Section~\ref{sec:limitations}): We discuss potential limitations encountered during the project and outline future research directions that could further enhance fairness in financial transactions.

\subsection{Key Findings}
\begin{figure}[ht]
   \centering
   \subfloat[]{\includegraphics[width=0.47\linewidth]{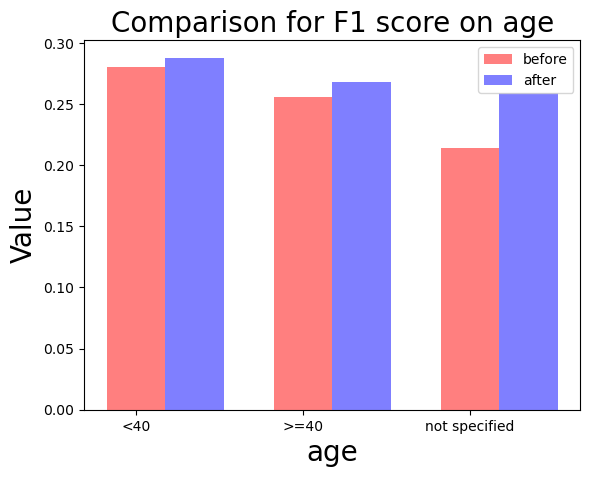}}
   \hfill
   \subfloat[]{\includegraphics[width=0.47\linewidth]{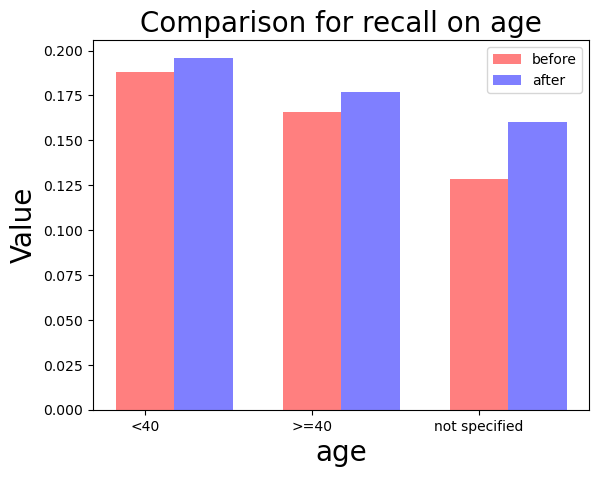}}
   \\[-2ex]
   \subfloat[]{\includegraphics[width=0.47\linewidth]{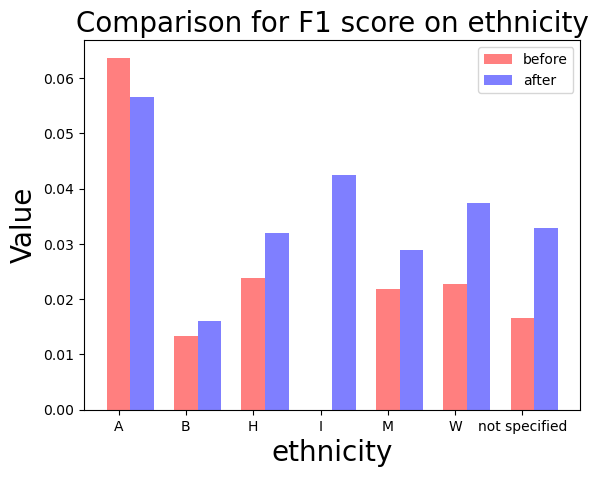}}
   \hfill
   \subfloat[]{\includegraphics[width=0.47\linewidth]{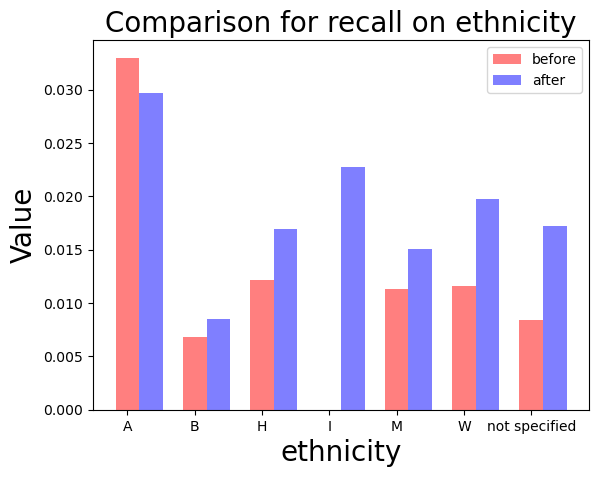}}
    \\[-2ex]
   \subfloat[]{\includegraphics[width=0.47\linewidth]{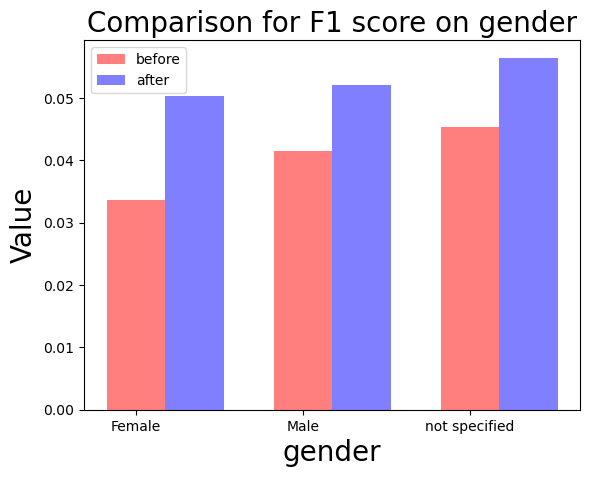}}
   \hfill
   \subfloat[]{\includegraphics[width=0.47\linewidth]{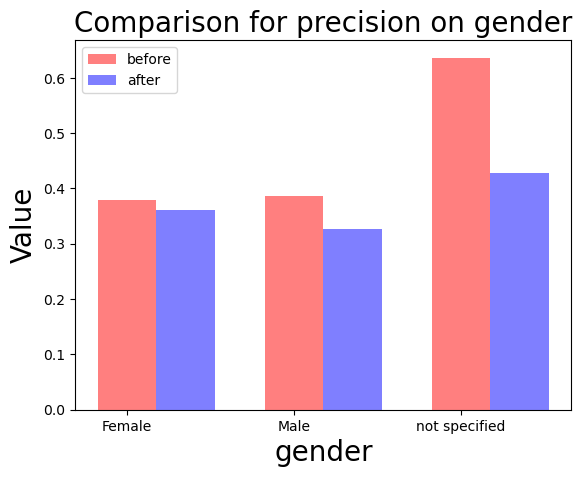}}
   \vspace*{-1.5mm}
   \caption{Model performance before and after applying Fairlearn on XGBoost classifier. The first row represents label 2. The second row represents label 3.  The third row represents label 7.}\label{fig: performance of fairlearn}
\end{figure}

\begin{itemize}
    \item \textbf{Intersectional Multi-Label Fairness} The analysis uncovered intersectional multi-label (industries) fairness challenges, identifying biases towards specific demographic groups in the data. The intersectionality of protected attributes may accentuate these biases.

    \item \textbf{Approach to Detecting Bias} In Section~\ref{sec:measuring} we present evidence that even when protected attributes are excluded from the training data, they still exhibit strong dependencies with other features in the dataset. These preliminary findings highlight the need to consider additional mitigation strategies, particularly for features that are proxy of these attributes.

    \item \textbf{Bias Mitigation:} In Section~\ref{sec:mitigation}, the results show that while certain mitigation strategies improve specific metrics, they may simultaneously increase bias in other areas. For example, compare Figure~\ref{miti_aequitas} (a) and (b): although overall fairness improves after mitigation, some groups still experience relatively worse outcomes. These experimental findings align with theoretical fairness impossibility theorems~\citep{miconi2017impossibility,DBLP:conf/nips/HsuMN022}. In addition, there are fairness-accuracy trade-off to take into consideration when mitigating for bias. As a consequence, stakeholders should assess trade-off across various protected attributes and labels, and weigh performance degradation against the benefits of fairer outcomes across the AI system. More research is needed to develop effective bias mitigation techniques, in particular for more complex intersectional, multi-label use cases.

    \item \textbf{Disparities across Industries}. The analysis revealed significant disparities in how different demographic groups fare across various industries. Some groups may dominate in one industry in terms of transaction amount while being less active in others (see Figure~\ref{fig: performance of fairlearn} for an example). This variability highlights the need for industry-specific fairness strategies to ensure fairness criteria are met across all sectors.

\end{itemize}

\subsection{Recommendations and Future Work}

\begin{itemize}

    \item \textbf{Stakeholder Engagement}: Managing high-dimensional sensitive attributes and multi-label industries requires decisions that align with specific business objectives and priorities. For example, in Figure~\ref{fig:3d_tensor_plot}, we observe a tensor representing a generalized approach to the intersectional and multi-label fairness problem. To optimize this representation and achieve an aggregate measure of fairness across these multiple dimensions, we recommend making decisions regarding the weighting or ranking for each dimension: protected attributes, industry labels, and metrics. These decisions may be made in collaboration with the different stakeholders of the model.


    \item \textbf{More Aligned Fairness Measures} Given the variety of fairness notions and metrics, We recommend to first identify which fairness concept is most appropriate for a given financial case study. This would help ensure that fairness metrics are aligned with business needs and societal values. 
    
    \item \textbf{Fairness-Accuracy Trade-Offs}: Implementing fairness mitigation techniques often reduces performance metrics. The first step is to assess whether improving model fairness without compromising model performance is possible. If this is not achievable, the critical question becomes: \textit{How much performance degradation is acceptable in order to optimize fairness?} This also requires a careful consideration of business needs and societal values. We recommend the use of Pareto frontiers to guide this decision.~\citep{DBLP:conf/aaai/ChiappaJSPJA20,DBLP:journals/jmlr/XuS23}

    \item \textbf{Usage of Sensitive Features during the Modeling.} Mastercard does not use demographic data as part of the models' training set. This approach is known as \enquote{fairness through unawareness}~\citep{dwork2012fairness,DBLP:conf/aies/SimonsBW21,DBLP:journals/ipm/CornacchiaABNPR23,DBLP:conf/fat/ChenKMSU19,DBLP:journals/jair/FabrisEMS23}. We have provided preliminary evidence that if fairness is understood as equality in the quality of service, algorithms that use the sensitive attribute as input may improve fairness (see Table~\ref{tab:metris_diff} in Section~\ref{sec:mitigation} for examples). This question requires more research from a technical, policy, and societal perspective.

    \item \textbf{Procedural Monitoring} We recommend to establish mechanisms to evaluate the entire model development pipeline for fairness, recognizing that the process by which the AI system is developed and makes decisions is often as important, if not more so, than the decisions themselves.

\end{itemize}

\subsection{Conclusion}

This report, resulting from a Data Study Group organized by the Alan Turing Institute in May 2024, encapsulates our interdisciplinary discussions and the empirical findings regarding fairness in financial transaction systems provided by Mastercard. Our analysis highlights the complexity of applying theoretical fairness concepts in real-world applications. It points to the necessity for novel methods and the adaptation of AI systems to reflect current societal values and technological advancements.

Moving forward, we suggest that Mastercard stakeholders continue working closely with machine learning engineers, data scientists, and the broader AI community to further refine the fairness assessment of their AI/ML models, and ensure that they perform effectively across various population groups. This effort requires sustained engagement from researchers, practitioners, and policymakers alike to ensure that the benefits of machine learning are accessible and fair for all.

\subsection{Disclaimer}
The data utilized in this study is synthetic and does not correspond to any real individuals or entities. Any resemblance to actual data is purely coincidental and unintended.

\section{Background: Fairness in Financial Transactions}\label{sec:background}

\subsection{Financial Transactions Models at Mastercard}
Mastercard is a global financial services company that provides payment processing products and services, facilitating electronic funds transfers worldwide with its branded credit, debit, and prepaid cards.

Mastercard develops predictive models that help issuing banks optimize card usage and prevent attrition. These models analyze patterns in usage, such as cross-border transactions and industry-specific spending, enabling banks to tailor their portfolios and offer targeted loyalty incentives to cardholders. As input these models use Mastercard’s data assets and expertise and allow issuing banks to identify and increase the revenue opportunities within their portfolios. For instance, they may feed a client’s lifecycle marketing efforts with account-level recommendations that will help them drive engagement, wallet share, and overall card spending.  Examples of types of predictions in this application include:

\begin{description}
    \item[Multi-label classification] predicts the likelihood of a cardholder's first-time spending in various industry-specific merchant categories over the next three months.

    \item [Multi-label regression] Estimates future spending amounts across different industry categories within the next three months.

\end{description}

This challenge aims to define, measure, and mitigate bias for Mastercard's financial transactions ML model. One of the main difficulties consists in defining fairness. There is a theoretical and contextual challenge in defining what constitutes fairness when predicting future spending across various industries based on historical patterns. Fairness can have different interpretations depending on the context and stakeholders involved, making it a complex concept to define accurately.



\subsection{Understanding Bias in Machine Learning Models}
Bias can lead to unfair treatment and disparities in the quality of service and allocation of resources. Here are the main types of bias that can affect machine learning models:

\begin{enumerate}
    \item \textbf{Representation/Selection Bias}: This occurs when the data used to train the model is not representative of the entire population. As a result, the model may perform well on the training data but poorly on new, unseen data, especially for underrepresented groups. We will see from our analysis that the dataset for this challenge inherently has this problem.~\citep{baeza2018bias}

    \item \textbf{Historical Bias}:  Historical bias denotes the entrenched biases that have developed over time within society, encompassing systemic inequalities, prejudices, and disparities. These biases become ingrained in various aspects of society, including the collection of data. As a result, data collection methods often mirror and perpetuate these preexisting biases, leading to skewed representations of certain groups or phenomena.

    \item \textbf{Measurement Bias}: Measurement bias arises when the features used in the model are measured inaccurately or inconsistently across different groups. This can lead to incorrect predictions and unfair treatment of certain groups.

    \item \textbf{Algorithmic Bias}: Algorithmic bias occurs when the model itself produces biased outcomes, often due to the way the algorithm is designed or optimized. This can happen even if the training data is unbiased~\citep{danks2017algorithmic}. In this study, we will try to mitigate this type of bias by introducing some in-processing and post-processing in section~\ref{sec:mitigation}.

    \item \textbf{Ground Truth Bias}: Ground truth bias happens when the labels ($Y$) used to train the model are biased. For example, if the training data labels reflect historical prejudices or inequalities, the model may learn and perpetuate these biases.  

    \item \textbf{Confirmation Bias}: This type of bias occurs when the model reinforces existing biases present in the data. If the model is trained on biased data, it may produce results that confirm these biases, creating a feedback loop of unfair outcomes.

    \item \textbf{Simpson’s paradox}: Simpson’s paradox can bias the analysis of heterogeneous data that consists of sub-groups or individuals with different behavioural patterns. According to Simpson’s paradox, a trend, association, or characteristic observed in underlying subgroups may be quite different from one subgroup to another~\citep{blyth1972simpson,DBLP:conf/aaai/Ruggieri0PST23}

    \item \textbf{Evaluation Bias}. Evaluation bias happens during model evaluation. It includes the use of either disproportionate or inappropriate benchmarks for the evaluation of applications. For instance, two common benchmark data sets used to test facial recognition systems have been shown to be biased towards skin colour and gender~\citep{buolamwini2018gender}.

\end{enumerate}

\subsection{Potential Harms in Machine Learning Models}

General machine learning models can have significant adverse effects on people. This can result in unfair behaviour, which we can define based on the harm or impact it has on individuals. We describe two different types of harms:~\citep{boyd2012critical,crawford2016there,crawford2017trouble}: 
\begin{itemize}

    \item \textbf{Allocation harms} may occur when AI models recommend allocation or withhold certain groups  opportunities, resources, or information. While the main applications are in sectors such as hiring or lending, allocation harms may occur in the marketing context  when, for instance, offers are not provided at the same rate to different groups  (“disparate promotion”). Examples include recommending a suboptimal card to one group or failing to offer promotions to a specific group.

     \item \textbf{Quality-of-service harms} may occur when a system does not perform equally for two groups, even if no opportunities, resources, or information are allocated or withheld. This is particularly relevant in a marketing context, for instance in product recommendation, when product recommendation may be ill-adapted to a specific group. In such case, the product recommendation for that group is of a lower quality, compared to other groups, resulting in quality-of-service harms.
\end{itemize}

\subsection{Existing Metrics for Measuring Fairness}

This work employs a specific subset of fairness metrics; for a comprehensive overview, refer to \cite{barocas-hardt-narayanan}.

Let us define:
\begin{itemize}
    \item $\mathbf{\hat{Y}} = (\hat{Y}_1, \hat{Y}_2, \ldots, \hat{Y}_L)$ as the vector of predicted labels for $L$ possible labels.
    \item $\mathbf{Y} = (Y_1, Y_2, \ldots, Y_L)$ as the vector of true labels.
    \item $A$ as the protected attribute.
    \item $\hat{Y}_i$ and $Y_i$ as the predicted and true label for the $i$-th label, respectively (where $\hat{Y}_i, Y_i \in \{0, 1\}$).
\end{itemize}

\subsubsection*{False Positive Parity}This metric ensures that the rate at which individuals are incorrectly classified as positive (false positives) is consistent across different groups, avoiding bias where one group experiences more false positive than another. False positive parity for the $i$-th label requires that the rate of false positives for that label is the same across all groups defined by the protected attribute $A$:
\[
P(\hat{Y}_i = 1 \mid Y_i = 0, A = a) = P(\hat{Y}_i = 1 \mid Y_i = 0, A = b)
\]
for all $a, b \in A$ and for all $i \in \{1, 2, \ldots, L\}$.

\subsubsection*{False Negative Parity} This metric guarantees that the rate at which individuals are incorrectly classified as negative (false negatives) is equal across groups, preventing a situation where one group is unfairly denied a positive outcome more often than others.
False negative parity for the $i$-th label requires that the rate of false negatives for that label is the same across all groups:
\[
P(\hat{Y}_i = 0 \mid Y_i = 1, A = a) = P(\hat{Y}_i = 0 \mid Y_i = 1, A = b)
\]
for all $a, b \in A$ and for all $i \in \{1, 2, \ldots, L\}$.

\subsubsection*{Overall Error Rate Parity}This metric ensures that the overall error rate, including both false positives and false negatives, is uniform across groups, promoting fairness by avoiding any group being systematically more misclassified than others.
Overall error rate parity for the $i$-th label requires that the total error rate (considering both false positives and false negatives) is the same across all groups:
\[
P(\hat{Y}_i \neq Y_i \mid A = a) = P(\hat{Y}_i \neq Y_i \mid A = b)
\]
for all $a, b \in A$ and for all $i \in \{1, 2, \ldots, L\}$.

\subsubsection*{Demographic Parity (DP)}Demographic parity aims for fairness by ensuring that the likelihood of a positive outcome does not depend on the protected attribute, thereby preventing any group from being favored or disfavored based on that attribute.
Demographic parity for the $i$-th label requires that the predicted outcomes are independent of the protected attribute:
\[
P(\hat{Y}_i = 1 \mid A = a) = P(\hat{Y}_i = 1 \mid A = b)
\]
for all $a, b \in A$ and for all $i \in \{1, 2, \ldots, L\}$.

\subsubsection*{Equality of Opportunity (EO)} This metric ensures that the true positive rate is the same across groups, meaning that if someone is actually positive, their chance of being correctly identified as positive is equal, regardless of their group.
Equality of opportunity for the $i$-th label is satisfied when the true positive rates are equal across different groups:
\[
P(\hat{Y}_i = 1 \mid Y_i = 1, A = a) = P(\hat{Y}_i = 1 \mid Y_i = 1, A = b)
\]
for all $a, b \in A$ and for all $i \in \{1, 2, \ldots, L\}$.

\subsubsection*{Conditional Use Accuracy Equality (CUAE)}
This metric ensures that both the positive predictive value (precision) and negative predictive value are consistent across groups, ensuring that the reliability of predictions is equal for everyone, regardless of their group.
This metric asserts that both positive predictive value and negative predictive value should be equal across groups:
\[
P(Y_i = 1 \mid \hat{Y}_i = 1, A = a) = P(Y_i = 1 \mid \hat{Y}_i = 1, A = b)\] \\
\[\text{and} \]
\[P(Y_i = 0 \mid \hat{Y}_i = 0, A = a) = P(Y_i = 0 \mid \hat{Y}_i = 0, A = b)
\]
for all $a, b \in A$ and for all $i \in \{1, 2, \ldots, L\}$.

\subsubsection*{Calibration by Group}  This metric ensures that for any given predicted probability, the actual chance of the outcome being positive is the same across groups, meaning that the predictions are equally trustworthy no matter which group an individual belongs to.
Calibration by group ensures that, for all predicted probabilities, the likelihood of the true label being positive is equal across groups:
\[
P(Y_i = 1 \mid \hat{P}_i = p, A = a) = P(Y_i = 1 \mid \hat{P}_i = p, A = b)
\]
where $\hat{P}_i$ is the predicted probability of the positive class for the $i$-th label, for all $a, b \in A$ and for all $p \in [0, 1]$.

\subsubsection*{Fairness as Protected Attribute Independence} A more restrictive definition of fairness can focus on the outcome, ensuring that decisions based on AI model recommendations are independent of sensitive features like race, gender, or age, across all label groups, including both privileged and underprivileged groups.

\subsection{Bias Mitigation Strategies}\label{preprocessors}
Efforts to mitigate unfairness in machine learning models can be broadly classified into three principal strategies \citep{DBLP:journals/widm/NtoutsiFGINVRTP20,DBLP:journals/ethicsit/AlvarezCEFFFGMPLRSSZR24,barocas-hardt-narayanan}:

\begin{itemize}
    \item \textbf{Preprocessing:} This strategy involves adjusting the input data  $\mathcal{D}$ to reduce bias before the data is utilized in the modelling process. Specifically, the sensitive attributes $A$ and other features $X$ are transformed so that the bias inherent in the data is minimized while retaining the essential statistical relationships. This could involve re-weighting data instances, synthesizing new data points, or altering feature values to enhance fairness. Literature examples~\citep{DBLP:conf/aistats/WardZC24,DBLP:conf/aies/MouganCRS23,zemel2013learning,DBLP:conf/aies/ZhangLM18,kamiran2012data}

    \item \textbf{In-Processing:} This method embeds fairness directly into the model training process. It modifies the learning algorithm to optimize both for prediction accuracy and fairness. This is typically achieved by defining a fairness-aware loss function:
    \[
    \underset{\theta}{\min} \, (\text{loss} + \lambda \cdot \text{fairness})
    \]
    where $\theta$ denotes the model parameters, $\text{loss}$ represents the standard loss function (e.g., mean squared error, cross-entropy), and $\lambda \cdot \text{fairness}$ is a regularization term that penalizes the model for unfair outcomes, with $\lambda$ controlling the trade-off between the original loss and the fairness objective. Literature examples~\cite{DBLP:journals/tkdd/WanZLZ23,agarwal2018reductions,mehrabi2021fairness,zafar2017fairness}

    \item \textbf{Post-Processing:} This approach modifies the outputs of an already trained model to enforce fairness. After the model, denoted as $h: (A, X) \rightarrow \hat{Y}$, generates its predictions $\hat{Y}$, these predictions are then adjusted to satisfy fairness criteria, potentially at the expense of some loss in accuracy. Adjustments can include altering decision thresholds for different groups or recalibrating the model’s output probabilities to equalize outcomes across groups. Literature examples~\cite{DBLP:journals/natmi/RodolfaLG21,DBLP:conf/nips/PetersenMSY21,DBLP:conf/aies/KimGZ19}
\end{itemize}

Each of these strategies has its strengths and is suitable for different stages of the model development lifecycle, depending on the specific fairness goals and constraints of the application.

\section{Data overview}\label{sec:data}

\subsection{Dataset description}

Mastercard provided participants with access to synthetic datasets based on financial transactions by customers in a 12-month period. The dataset consists of 1 million rows, 20 features, and 9 output labels. Each row describes the spending habits of every customer in different industry categories. Finally, for every customer, we have additional information on demographics described by three attributes that can be a source of a potential multidimensional discrimination~\citep{DBLP:conf/fat/0001HN23}. Note that this demographic data comes from a third party datasource, as Mastercard does not collect demographic data on cardholders:
\begin{itemize}
    \item \textbf{Gender:} Female, male, or unspecified
    \item \textbf{Ethnicity:} White, Hispanic, Black, Asian, Middle Eastern, Native Indian, or not given/unspecified.
    \item \textbf{Age:} Below or above 40 years old
\end{itemize}

We have two datasets:

\begin{itemize}
    \item \textbf{Adoption} describes the binary value of whether or not the customer will spend within a particular category in the next 3 months.

    \item \textbf{Spending} consists of labels specifying the amount of spending by the customer in a particular industry category in the next 3 months.
\end{itemize}

\subsection{Data quality}

Following ~\cite{DBLP:conf/nips/MouganPTBECJSWM23}  general data quality guidelines in datathons, we considered the following aspects of data quality:

\begin{description}

    \item[Appropriateness]\emph{Was the data relevant to the questions posed?}\\
    The data was relevant. The focus of this research was on the methodological approach to bias. The data enabled empirical testing of the proposed methodological approach for bias assessment.
    
    \item[Readiness]\emph{ Is the data complete, is relevant metadata available, and is there sufficient data documentation?}\\

    Yes, the data was documented and Mastercard provided a data dictionary.

    \item[Reliability / Bias] \emph{To what extent do the data accurately represent the population or phenomenon? Are there systematic data collection errors or distortions?}\\
    
    It is worth noting that approximately 30\% of the data in the ethnicity column is unknown because that data was not available. Since Mastercard does not collect demographic data on its cardholders, this information is provided by a Third-Party data source. In the case of ethnicity, this variable obtained from the Third-Party dataset is modelled, which might explain the large number of missing values for ethnicity (cf. Figure~\ref{fig:eda.ethnicity}). In contrast, the other demographics, such as age (only 0.02\%) and gender (3\%), have a relatively lower percentage of undisclosed entries (NaN). Furthermore, upon removing the unspecified data, Caucasians comprise 70\% of the ethnicity column, leaving the remaining 30\% to represent all other demographic groups. This discrepancy indicates both overrepresentation and underrepresentation of different ethnicities within the dataset compared to the overall US population as captured in the US census data.
    
    \item[Sensitivity]\emph{Was the data private or confidential?} \\
    The data is synthetic, meaning that it describes the behavior of made-up individuals, but in a way that still reflects the trends and patterns in the original dataset.

    \item[Sufficiency]\emph{Was the amount of data enough?}\\
    
    We found that data had a large class imbalance in terms of positive and negative examples. Thus, for sufficient learning, a larger sample of positive examples would be needed. 

\end{description}

\subsection{Exploratory Data Analysis}

\subsubsection{Labels}

In our exploratory data analysis for the labels, we found two main issues that need to be addressed jointly. First, the distribution of labels is noticeably skewed, which may amplify performance disparity in predictive modelling. This skewness can lead to models that perform well for the majority labels while neglecting the minority labels. Second, demographic disparities are widely different across labels. This insight suggests that the notion of privileged groups is not the same across all labels and that the intersectionality of different demographic groups is a key factor in understanding the distribution of harm.

\begin{figure}
    \centering
    \includegraphics[width=1\linewidth]{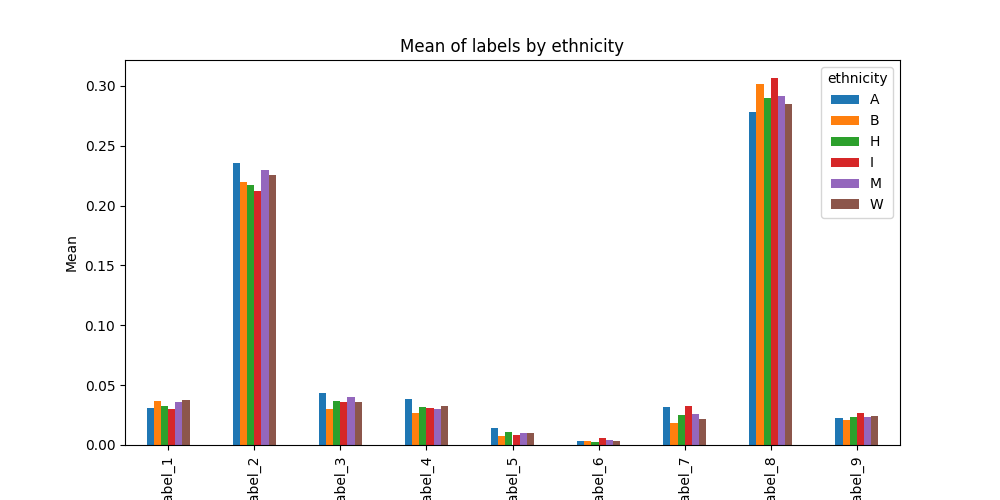}
    \caption{Labels imbalance in adoption dataset. Labels 2 and 8 have disproportionately more positive instances.}\label{fig:labels}
\end{figure}

In terms of label specifics, label 6 represents the highest spending industry but is paradoxically the least adopted, implying a concentration of spending in a small fraction of the population (cf. Figure~\ref{fig:labels}). In contrast, labels 2 and 8 show a consistent pattern of the highest spending coupled with the highest adoption rates, suggesting that not all labels (industries) matter equally when looking at model performance and business objectives. A minor yet noteworthy finding is the negative correlation (approximated at -0.3) among the two dominant labels, 2 and 8. This insight suggests that labels should be considered jointly in predictive modelling, as there is underlying information on the joint distribution of these labels.
In addition, this highlights the fact that this multi-label dataset is highly imbalanced (with important disparities on spending amount and adoption across industries) which adds further complexity to the bias testing of a multi-label spending model. 

\FloatBarrier
\FloatBarrier

\FloatBarrier
\begin{figure}
    \centering
    \includegraphics[width=1\linewidth]{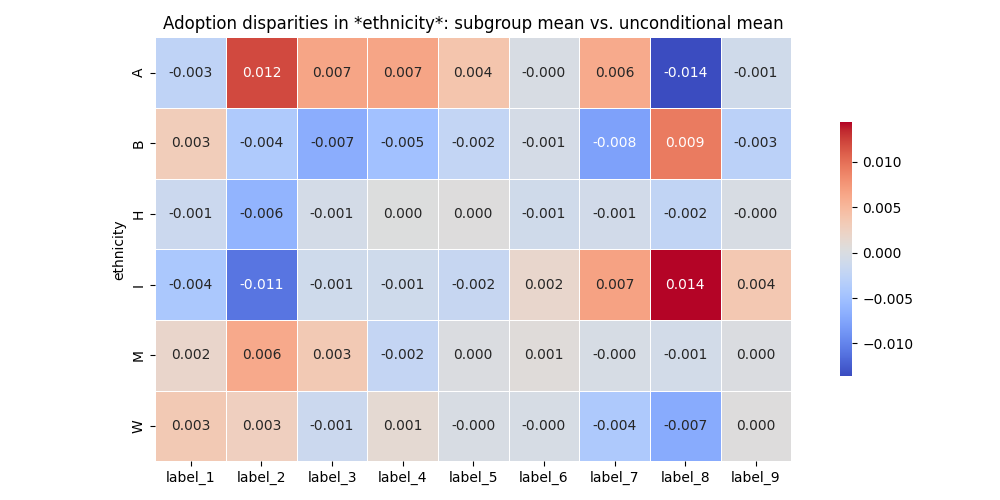}
    \caption{Heterogeneous disparities within the label for ethnicity in adoption dataset. Each value represents how much each subgroup's mean differs from the unconditional mean within a given label.}
\end{figure}\label{fig:eda.ethnicity}
\FloatBarrier

\subsubsection{Features}

On the feature side, our analysis revealed that most features in the data have a similar bimodal distribution, clustering at two extremes (minimum and maximum values), with very little density in the middle. Furthermore, when we compare the distributions of features across two sub-populations: those who have all zeros labels and those who have at least one non-zero label, we find that lower density at both extremes tends to correlate with higher spending and adoption rates. This suggests that the most relevant information lies in the middle of the feature distribution, where there are almost zero density.

We also noticed that some features strongly correlate with each other, usually appearing in pairs (cf. Figure~\ref{fig:correlat}). This could be due to the synthetic nature of the data, where some features are statistics based on the same underlying feature. Together with the previous insight, this suggests that downstream tasks such as multi-label classification and regression would have poor performance, as the data itself has strong limitations regarding information content.

\FloatBarrier
\begin{figure}
    \centering
    \includegraphics[width=1\linewidth]{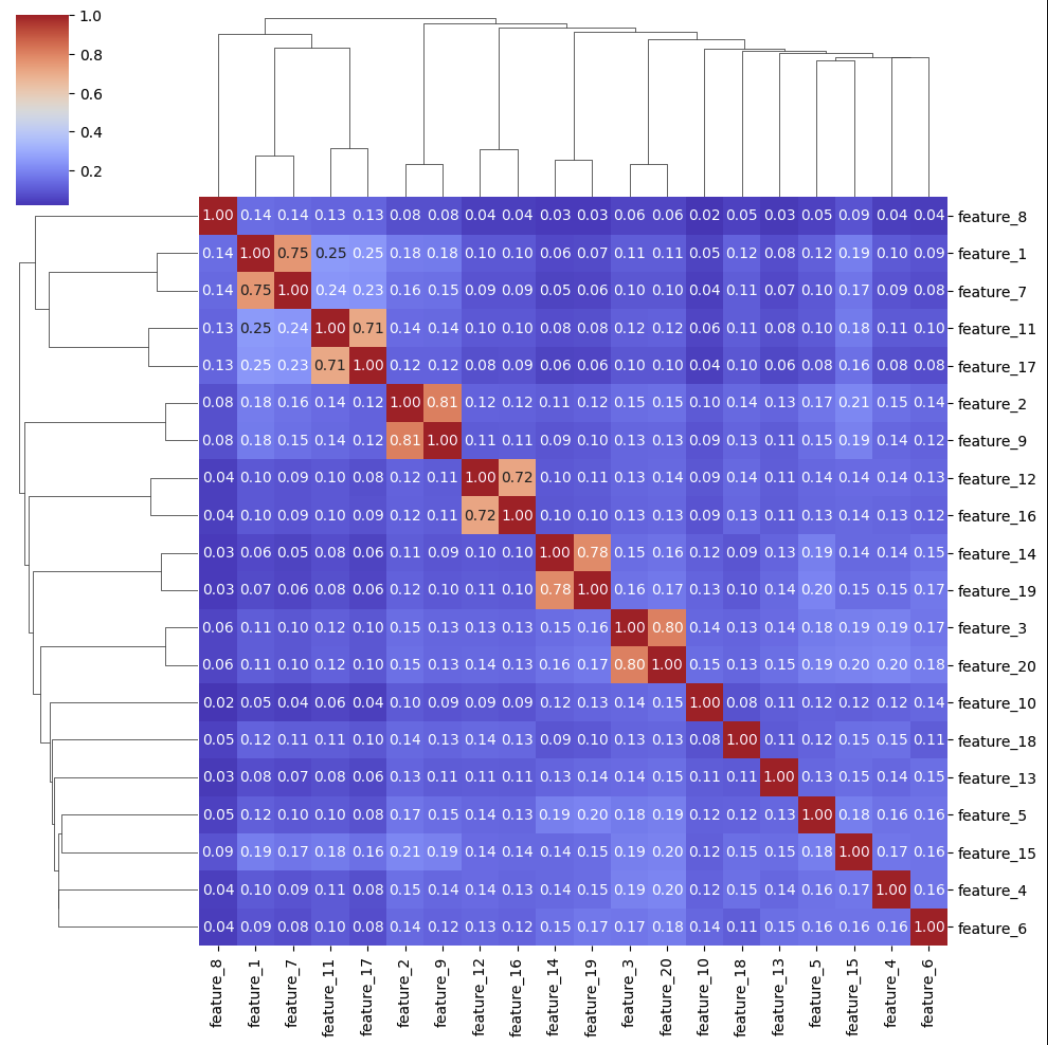}
    \caption{Features correlation in spending data}
\end{figure}\label{fig:correlat}
\FloatBarrier

\section{Mathematical Formulation}\label{sec:math.formulation}
In this section, we will provide the formal notations that we utilize throughout this study.
\begin{table}[ht]
\centering
\begin{tabular}{|c|l|}
\hline
\textbf{Notation} & \textbf{Description} \\ \hline
$A$ & Sensitive attribute \\ \hline
$X$ & Covariates \\ \hline
$Y$ & Target variable \\ \hline
$\mathbf{X} = \{X_1, \ldots, X_p\}$ & Set of $p$ predictive features \\ \hline
$\mathcal{D}$ & Dataset consisting of $\{A, X, Y\}$ \\ \hline
$f_\theta: \mathbf{X} \to Y$ & Model function \\ \hline
$\hat{Y}_i$ & Predicted label for the $i$-th instance \\ \hline
$Y_i$ & True labels for the $i$-th instance (row) \\ \hline
$Y^k$ & True labels for the $k$ class for all instances column) \\ \hline
$y^k_i$ & True labels for the $k$ class $i$-th instance \\ \hline
$\theta$ & Model's parameters \\ \hline
$\mathit{dom}(Y)=\{0, 1,\ldots, k\}$ & Domain of the target feature in a multi-label task \\ \hline
$y^k_i$ & $k$-class of the $i$-th instance \\ \hline

$\D_{\mathbf{X}}{} = \{\mathbf{x}_1, \ldots, \mathbf{x}_n\}$ & Empirical input distribution \\ \hline
$\mathbf{A}=\{A_0,A_1,\ldots,A_9\}$ & Range of values for the protected feature \\ \hline
\end{tabular}
\caption{Notation Table for Supervised Learning Model}
\label{tab:notation}
\end{table}

A supervised learning model is a function $f_\theta: \mathbf{X} \to Y$ induced from a set of observations, called the training set, where $\mathbf{X} = \{X_1, \ldots, X_p\}$ are $p$ predictive features, $Y$ is the target feature, and $\theta$ are the models' parameters.

The domain of the target feature is multi-label task $\mathit{dom}(Y)=\{0, 1,\ldots k\}$,  and the $k$-class of $i$-the instance can be denoted as $y^k_i$.

We assume a probabilistic estimator, and we denote by $f_\theta(\mathbf{x})$ that estimates the probability $P(Y=1|\mathbf{X}=\mathbf{x})$ over the (unknown)  distribution of $\mathbf{X} \times Y$. For regression, $f_\theta(\mathbf{x})$ estimates $E[Y|\mathbf{X}=\mathbf{x}]$.

We assume we have a dataset $\mathcal{D}$ consisting of $\{A, X, Y\}$, which represent the sensitive attribute, covariates, and target variables, respectively.

We assume a feature $A$ representing social groups, called the \textit{protected feature}, which can take  values in the range$\mathbf{A}=\{A_0,A_1,\ldots,A_9\}$. It is modelling membership to a protected social group. $Z$ can either be included in the predictive features $\mathbf{X}$ or not. If not, we assume that it is still available.

\subsection{Multi-Class Protected Attribute and Multi-Label Fairness}

For multi-label classification, let $\mathit{dom}(Y) = \{0, 1\}^K$, where each label $\mathbf{y}_i = (y_i^1, y_i^2, \ldots, y_i^K)$ is a vector of binary values indicating the presence or absence of each class. The prediction for instance $i$ is denoted as $\hat{\mathbf{y}}_i = (\hat{\mathbf{y}}_i^1, \hat{\mathbf{y}}_i^2, \ldots, \hat{\mathbf{y}}_i^K)$. In our case, we have $K=9$.

For protected attributes, let $\mathbf{A} = \{A_0, A_1, \ldots, A_L\}$. In our case, we have $L=24$ distinct protected attribute dimensions indicating intersectional discrimination, for example $A_0=\texttt{Male-Age>40-EthnicA}$

For each entry industry class and for each value of the protected attribute, we will have a vector representing the fairness comparison against each other protected attribute. For example, for $A_0$, we have:

\[
\mathbf{A}_0 = \{A_0 - A_1, A_0 - A_2, \ldots, A_0 - A_L\}
\]

with dimensions $L-1$.

Therefore,

If we define any fairness metric $g$ that receives as input a distribution $X$, labels $\mathbf{Y}_K$, and a protected attribute $A_l$, we obtain a tensor $G \in \mathbb{R}^{L \times K \times K}$ where:

\[
G_{l,k_1,k_2} = g(X, \mathbf{Y}_{k_1}, A_l) - g(X, \mathbf{Y}_{k_2}, A_l)
\]

\begin{figure}[ht]
    \centering
    \includegraphics[width=1\linewidth]{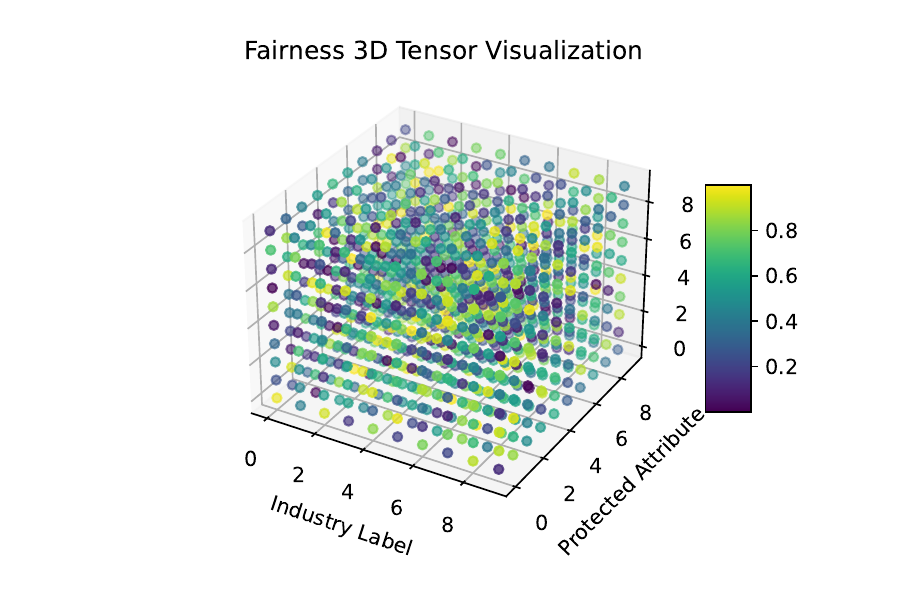}
    \caption{Visualization of the tensor that formalizes the problem of intersectional fairness in multilabel problem $G_{l,k_1,k_2} = g(X, \mathbf{Y}_{k_1}, A_l) - g(X, \mathbf{Y}_{k_2}, A_l)$. The scale represents any fairness metric $g\in\{0,1\}$}
    \label{fig:3d_tensor_plot}
\end{figure}

This tensor $G$ encapsulates the fairness comparisons across all protected attributes and label pairs(cf. Figure~\ref{fig:3d_tensor_plot}). This formalization opens two research questions that we will discuss on the next section.

\subsection{Incorporating Weights into the Fairness Tensor}

To enhance the adaptability of our fairness metrics in multilabel and intersectional fairness settings, we introduce a weighting scheme. This scheme accounts for the varying significance of different classes and protected attributes in our analysis.

\textbf{Weighted Tensor Definition}
Consider a tensor \( G \in \mathbb{R}^{L \times K \times K} \), defined in the previous section, which encapsulates fairness comparisons across all protected attributes and label pairs. We extend this tensor by incorporating a weight matrix \( W \), where \( W \in \mathbb{R}^{L \times K} \). Each element \( W_{l,k} \) represents the weight for the protected attribute \( l \) and the label class \( k \), signifying its importance or relevance in the fairness analysis.

\subsection{Formulation of the Weighted Fairness Tensor}
The weighted fairness tensor, \( G^W \), is then defined as follows:
\[
G^W_{l,k_1,k_2} = W_{l,k_1} \cdot G_{l,k_1,k_2} - W_{l,k_2} \cdot G_{l,k_1,k_2}
\]

This formulation modifies the original fairness comparisons in the tensor \( G \) by scaling each term by the corresponding weights in \( W \). This approach allows the model to emphasize or de-emphasize specific fairness concerns based on the dataset characteristics or policy requirements.

\textbf{Interpretation}
The elements of \( G^W \) now reflect weighted fairness measures. A higher weight \( W_{l,k} \) increases the influence of the fairness discrepancy between the labels \( k_1 \) and \( k_2 \) under the protected attribute \( l \). This is particularly useful in scenarios where certain disparities are deemed more critical than others.

\textbf{Application}
The weighted tensor \( G^W \) can be used to prioritize interventions and policy adjustments. By adjusting the weights in \( W \), stakeholders can tailor the fairness analysis to reflect specific equity goals or compliance requirements.

\section{Measuring Fairness}\label{sec:measuring}

This section is divided into two parts: First we assess the relationship of covariates with the protected attribute and then we study fairness in both multi-classification and multi-regression models.

\subsection{Measuring Proxy Discrimination by Predicting Protected Characteristics from Covariates}

\textbf{Motivation} The goal of this section is to measure potential proxy discrimination by evaluating whether machine learning models can predict protected characteristics (like age or gender) from other available data. This is done using a classifier two-sample test approach, where a machine learning classifier (in this case, XGBoost) is trained to distinguish between samples that differ in the protected characteristic.

\textbf{Methodology} By training the classifier and analyzing its performance (using metrics like accuracy, precision, recall, and F1-score), the researchers can determine how well the model can infer protected characteristics from other covariates, which may indicate proxy discrimination.  We can highlight the flexibility and interpretability offered by classifier two sample test using machine learning models for high-dimensional data.

The use of SHAP values further helps to interpret which features contribute most to the model's ability to predict the protected characteristic, offering insights into potential indirect dependencies. The results for age and gender prediction are presented to demonstrate the classifier's effectiveness, and recommendations are made for future research and ethical considerations.

\paragraph{Classifier Two-Sample Test on Protected Characteristics}

One of the fundamental problems in statistics is determining whether two samples, $\mathbf{S}_P \sim P^n$ and $\mathbf{S}_Q \sim Q^m$, originate from the same probability distribution. To address this, two-sample tests \citep{lehmann1986testing} summarize the differences between the two samples into a real-valued test statistic, which is then used to either accept or reject the null hypothesis that $P = Q$. Developing powerful two-sample tests is crucial for numerous applications, including the evaluation and comparison of generative models.

Over the past century, statisticians have developed a wide variety of two-sample tests. However, many of these tests are limited to one-dimensional data, require a fixed data representation, yield test statistics in units that are difficult to interpret, or fail to explain how the two samples differ. These limitations have spurred ongoing research to create more versatile and interpretable two-sample tests, especially for complex, high-dimensional data.

In recent years, machine learning (ML) classifiers have emerged as powerful tools for conducting two-sample tests, especially in high-dimensional and complex data settings. The basic idea is to train an ML classifier to distinguish between instances from the two samples, $\mathbf{S}_P \sim P^n$ and $\mathbf{S}_Q \sim Q^m$. If the classifier can effectively differentiate between the two samples, it suggests that the distributions $P$ and $Q$ are different. This test can serve as a way to measure relationship between variables in high dimensional and heterogeneous data sources, where classical statistical testing will have either false positives, false negatives  or lack explanation capabilities. 

The procedure typically involves the following steps:
\begin{enumerate}
    \item \textbf{Label Assignment:} Assign a binary label to each instance, where instances from $\mathbf{S}_P$ are labelled as 0 and instances from $\mathbf{S}_Q$ are labeled as 1.
    \item \textbf{Classifier Training:} Train an ML classifier, such as a logistic regression, support vector machine (SVM), or neural network, using the labelled instances. The classifier learns to predict the probability that an instance belongs to either $\mathbf{S}_P$ or $\mathbf{S}_Q$.
    \item \textbf{Evaluation:} Evaluate the classifier's performance using metrics such as accuracy, AUC-ROC, or precision-recall. A high performance indicates that the samples are distinguishable, implying that $P \neq Q$.
    \item \textbf{Test Statistic:} The classifier's performance metric can be used as the test statistic. For instance, the accuracy or the AUC-ROC score can serve as a measure of how well the classifier discriminates between the two samples.
    \item \textbf{Hypothesis Testing:} Compare the classifier's performance to a null distribution obtained through permutation testing. This involves repeatedly shuffling the labels of the combined dataset and retraining the classifier to establish a distribution of performance metrics under the null hypothesis that $P = Q$.
\end{enumerate}

Using an ML classifier for two-sample tests offers several advantages:
\begin{itemize}
    \item \textbf{Flexibility:} ML classifiers can handle high-dimensional data and complex relationships between features, making them suitable for a wide range of applications.
    \item \textbf{Interpretability:} Techniques such as feature importance scores and model explanations (e.g., SHAP values) can provide insights into which features contribute most to the differences between the samples.
    \item \textbf{Robustness:} ML classifiers are robust to various data distributions and can be adapted to different types of data, including structured, unstructured, and time-series data.
\end{itemize}

This approach has been successfully applied in various domains, including genomics, image analysis, and natural language processing, where traditional two-sample tests may fail due to the complexity and high dimensionality of the data\citep{DBLP:conf/iclr/Lopez-PazO17,chakravarti2023model,mougan2024beyond}.

\paragraph{Using XGBoost and SHAP} In the following section we use a \texttt{xgboost} as classifier and evaluate the model reporting the performance and its Shapley values~\cite{DBLP:conf/nips/LundbergL17,DBLP:journals/natmi/LundbergECDPNKH20}
\clearpage
\subsubsection{Predicting Age}
\begin{table}[ht]
    \centering
    \begin{tabular}{lcccc}
        \toprule
        \textbf{Metric} & \textbf{Class 0} & \textbf{Class 1} & \textbf{Macro Avg} & \textbf{Weighted Avg} \\
        \midrule
        Precision & 0.72 & 0.55 & 0.63 & 0.67 \\
        Recall & 1.00 & 0.01 & 0.50 & 0.72 \\
        F1-score & 0.84 & 0.02 & 0.43 & 0.61 \\
        Support & 143612 & 55611 & 199223 & 199223 \\
        \midrule
        \multicolumn{5}{l}{\textbf{Accuracy: 72.13\%}} \\
        \midrule
        \multicolumn{5}{l}{\textbf{Confusion Matrix:}} \\
        \midrule
        \multicolumn{5}{l}{
            \begin{tabular}{cc}
                143181 & 431 \\
                55092 & 519 \\
            \end{tabular}
        } \\
        \bottomrule
    \end{tabular}
    \caption{Classification Report}
    \label{tab:classification_report}
\end{table}

\begin{figure}
    \centering
    \includegraphics[width=1\linewidth]{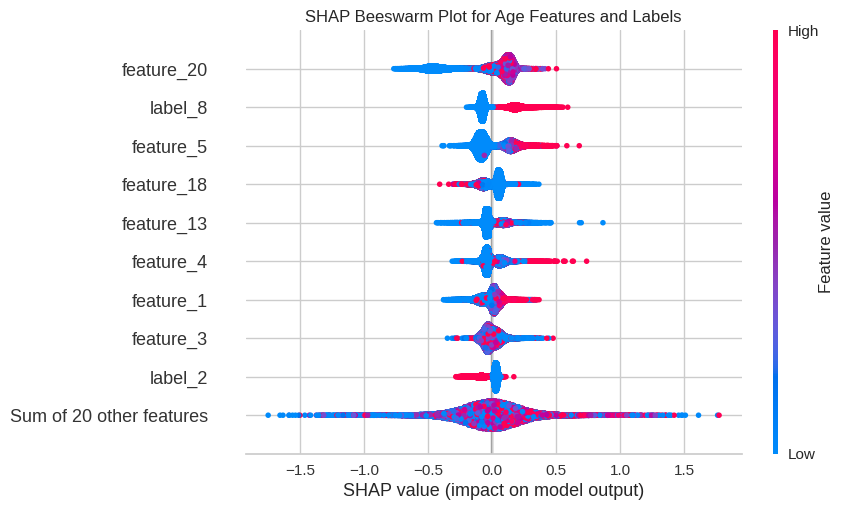}
    \caption{SHAP Beeswarm values for XGBoost model trained on features 1-20 and labels predicting age}\label{fig.shap.beeswarm}
\end{figure}

\begin{figure}
    \centering
    \includegraphics[width=1\linewidth]{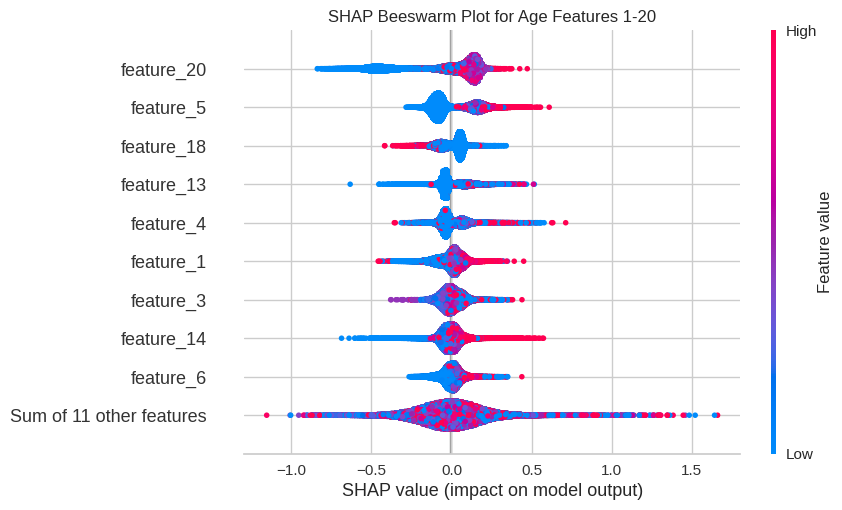}
    \caption{SHAP Beeswarm values for XGBoost model trained on features 1-20 predicting age}\label{fig:shap.beeswarm2}
\end{figure}

\clearpage
\subsubsection{Predicting Sex}
\begin{table}[ht]
    \centering
    \begin{tabular}{lcccc}
        \toprule
        \textbf{Metric} & \textbf{Class 0} & \textbf{Class 1} & \textbf{Macro Avg} & \textbf{Weighted Avg} \\
        \midrule
        Precision & 0.56 & 0.61 & 0.59 & 0.59 \\
        Recall & 0.40 & 0.75 & 0.58 & 0.60 \\
        F1-score & 0.47 & 0.67 & 0.57 & 0.58 \\
        Support & 85079 & 106446 & 191525 & 191525 \\
        \midrule
        \multicolumn{5}{l}{\textbf{Accuracy: 59.64\%}} \\
        \midrule
        \multicolumn{5}{l}{\textbf{Confusion Matrix:}} \\
        \midrule
        \multicolumn{5}{l}{
            \begin{tabular}{cc}
                34161 & 50918 \\
                26391 & 80055 \\
            \end{tabular}
        } \\
        \bottomrule
    \end{tabular}
    \caption{Classification Results XGBoost detecting Sex on features}
    \label{tab:classification_results}
\end{table}

\begin{figure}
    \centering
    \includegraphics[width=1\linewidth]{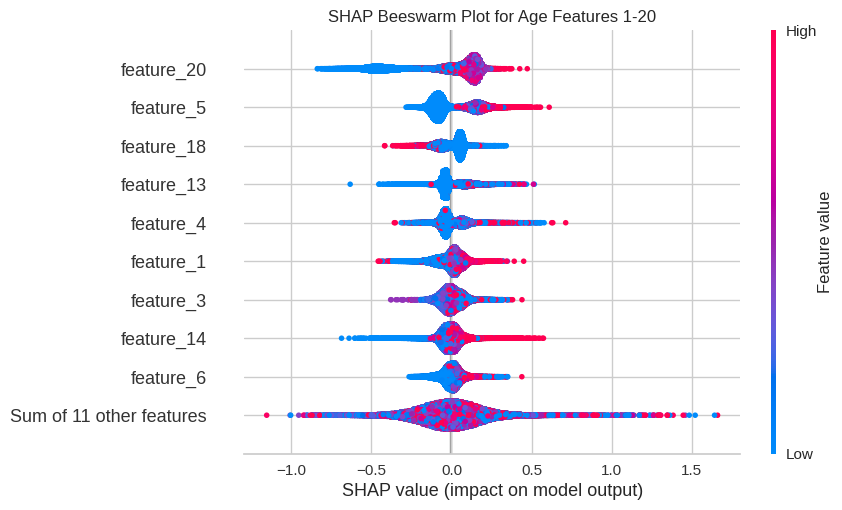}
    \caption{SHAP Beeswarm values for XGBoost model trained on features 1-20 predicting sex}\label{fig:shap.beeswarm3}
\end{figure}

\begin{figure}
    \centering
    \includegraphics[width=1\linewidth]{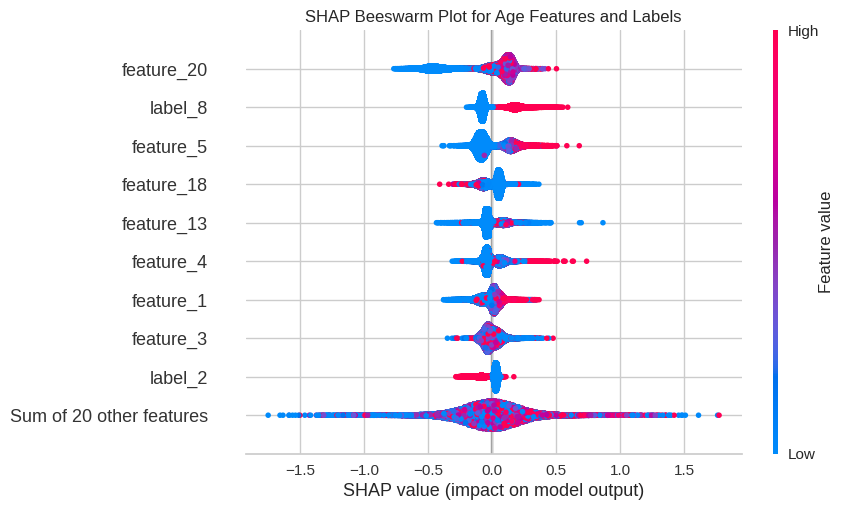}
    \caption{SHAP Beeswarm values for XGBoost model trained on features 1-20 and labels predicting sex}\label{fig:beeswarm4}
\end{figure}

\paragraph{Conclusion}

In this section, we explored the use of machine learning classifiers for measuring data dependencies with the protected attribute by conducting two-sample tests on protected characteristics, specifically focusing on predicting age and sex from data features. Traditional two-sample tests, while valuable, often face limitations when dealing with high-dimensional and complex data. We can overcome these limitations by leveraging ML classifier's two-sample test, offering greater flexibility, interpretability, and robustness.

The results demonstrated that even if the protected attribute is not present in training data, it has strong dependencies with other features on the dataset. For age prediction, the classifier achieved an accuracy of 72.13\%, while for sex prediction, the accuracy was 59.64\%(cf. Figure~\ref{fig.shap.beeswarm},Figure~\ref{fig:shap.beeswarm2},Figure~\ref{fig:shap.beeswarm3},Figure~\ref{fig:beeswarm4}). 

\textbf{Recommendation:} Future work could focus on refining these models, exploring other ML algorithms, and applying this approach to a broader range of protected characteristics and data types. Additionally, the ethical implications of using ML for predicting protected characteristics must be carefully considered to ensure fairness and avoid potential biases~\cite{kosinski2013private}.

\subsection{Multi-Label Classification} \label{multilabel_classification}

In measuring the fairness of our models, we first train different multi-label classification models and measure bias in terms of different metrics. In this section, we provide details of the baseline model and our best model performance in terms of the different metrics we considered in our study.

\subsubsection{Baseline Models (Logistic Regression Model)}
We use logistic regression as our baseline model for the multi-label classification problem~\citep{cho2023multi, tao2019novel}. Logistic regression is widely used for both binary and multi-class classification tasks. For our multi-label classification, we adopted the OneVsRestClassifier from sci-kit-learn. This approach enables us to handle multi-label classification by treating it as a series of binary classification problems, where a separate logistic regression model is trained independently for each label. This method leverages the simplicity and efficiency of logistic regression while extending its application to multi-label classification scenarios. We include in this report the details on the performance of the logistic regression model across the different demographic groups considered in this study.

\FloatBarrier
\begin{figure}[htbp]
\centering 
\includegraphics[width=1\textwidth]{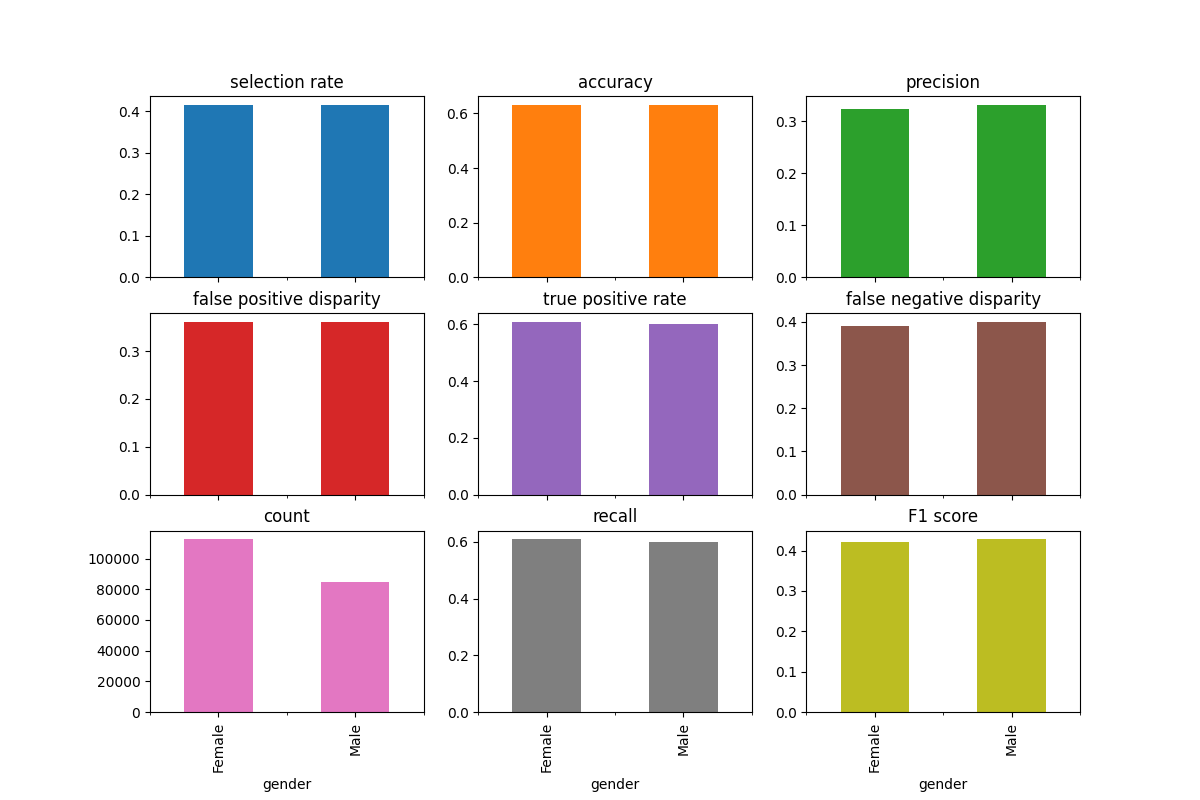}
\caption{Multi label classification model for label $2$}
\label{multi_label_2}
 \end{figure}
\FloatBarrier
The plot in figure~\ref{multi_label_2} displays the performance metrics of an unmitigated multi-label classification model and we show the results for label $2$, evaluated across different gender groups (i.e. female and male). The plot is divided into three rows, where we show different performance metrics:  selection rate, accuracy, precision, false positive disparity, true positive rate, false negative disparity, recall and $F1score$,
Each metric is presented for the different gender groups in our datasets. We further explain a few of the metrics for this plot below:
\begin{itemize}
\item \textbf{Accuracy:} Both gender group shows an equal accuracy, indicating that the models perform equally well for these demographic groups in terms of overall correct predictions. However, this accuracy is nothing better than a random classifier.

\item \textbf{True Positive Rate (TPR)}: The true positive rates are identical for both genders. This indicates that the models have an equal ability to identify positive instances for both female and male groups correctly.

\item \textbf{False Negative Rate (FNR)}: Similarly, the false negative rates are consistent across genders and models, suggesting that the likelihood of incorrectly predicting a negative instance is the same for both groups.
\end{itemize}
From the above model performance, we can conclude that the multi-classification shows an almost equal performance across the different genders, even though the classifier does not perform any better than a random classifier.

\subsubsection{XGboost}

We also adopted XGBoost (eXtreme Gradient Boosting)~\citep{chen2016xgboost} to efficiently and effectively implement the gradient boosting algorithm. It uses advanced optimisation techniques, such as tree pruning, and efficient memory usage to enhance training speed and model performance. We adopt a multi-output scheme to train our XGBoost model so that the model gives predictions of all nine target labels at the same time. 

To train and validate the model, we split the dataset into training and testing sets which consist of randomly sampled 70\% and 30\% of the data respectively. The distribution of positive and negative labels is consistent across training and testing sets. The hyperparameters of the model, including max\_depth and n\_estimators, have been optimised in a grid search using recall as a target metric. 

\paragraph{Results}

We evaluated the performance of the resulting model for each label separately using multiple classification metrics (such as accuracy, true positive rate, f1 score etc) in \prettyref{tab:metrics}. Given that the purpose of the classifier is to identify cardholders who are more likely to spend on each industry (label), we focus on recall (true positive rate) in the evaluation of the model's performance. It is worth noting that the model's performance is particularly higher for labels 2 and 8 compared to the rest of the labels, which is most likely due to the fact that the number of positive samples for the two labels is higher. 

\begin{table}[ht]
\centering
\resizebox{\textwidth}{!}{%
\begin{tabular}{cccccccc}
\toprule
label & selection rate & accuracy & precision & false positive rate & true positive rate & false negative rate  & F1 score \\
\midrule
1 & 0.016 & 0.95 & 0.18 & 0.013 & \textbf{0.078} & 0.92 & 0.11 \\
2 & 0.16 & 0.74 & 0.40 & 0.13 & \textbf{0.29} & 0.71 & 0.33 \\
3 & 0.016 & 0.95 & 0.17 & 0.014 & \textbf{0.074} & 0.93 & 0.10 \\
4 & 0.014 & 0.96 & 0.19 & 0.012 & \textbf{0.087} & 0.91 & 0.12 \\
5 & 0.0027 & 0.99 & 0.10 & 0.0025 & \textbf{0.031} & 0.97 & 0.048 \\
6 & 0.00052 & 1.00 & 0.019 & 0.00052 & \textbf{0.0032} & 1.00 & 0.0054 \\
7 & 0.0087 & 0.97 & 0.18 & 0.0073 & \textbf{0.070} & 0.93 & 0.10 \\
8 & 0.24 & 0.69 & 0.46 & 0.18 & \textbf{0.37} & 0.63 & 0.41 \\
9 & 0.0097 & 0.97 & 0.18 & 0.0082 & \textbf{0.074} & 0.93 & 0.10 \\
\bottomrule
\end{tabular}
}
\caption{Performance of multilabel XGBoost classifier by labels}
\label{tab:metrics}
\end{table}

\begin{figure}[ht]
    \centering
        \centering
        \includegraphics[width=1\textwidth]{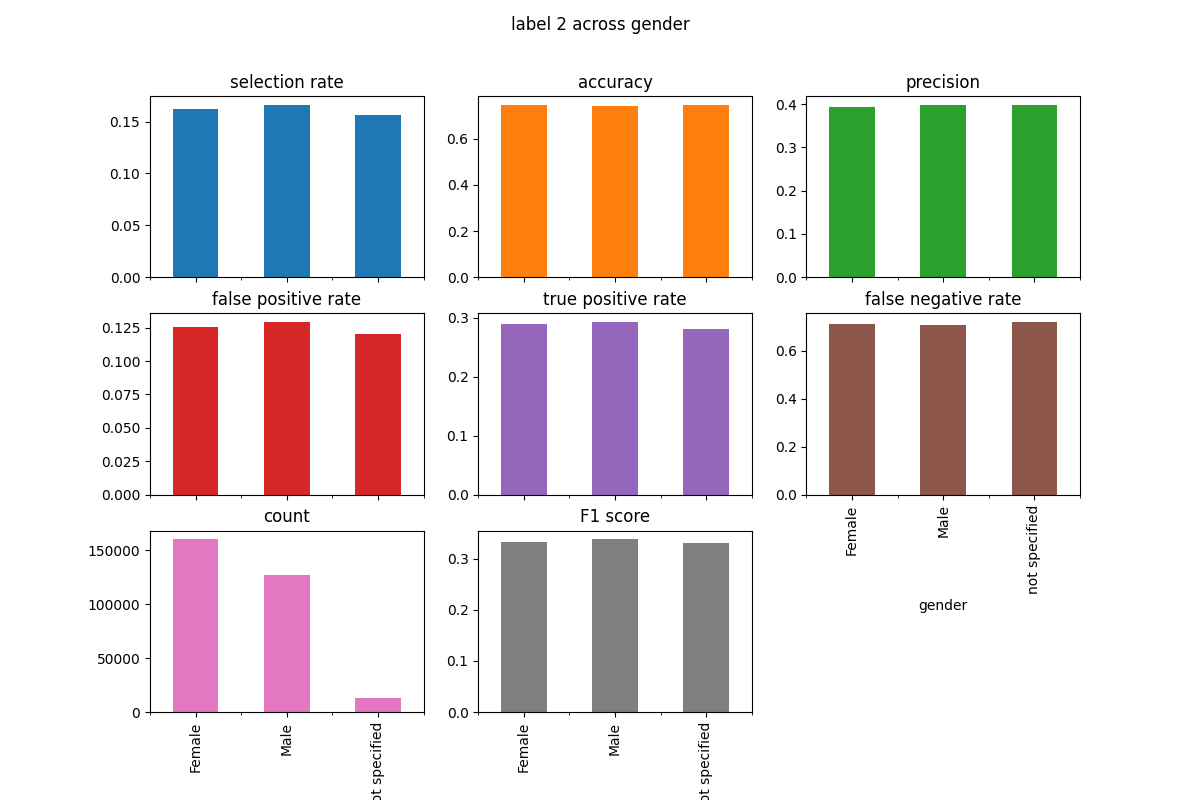}
        \caption{XGBoost classifier performance for label 2 by gender}
        \label{fig:xgb_gender}

\end{figure}

\begin{figure}[ht]
    \centering
        \centering
        \includegraphics[width=1\textwidth]{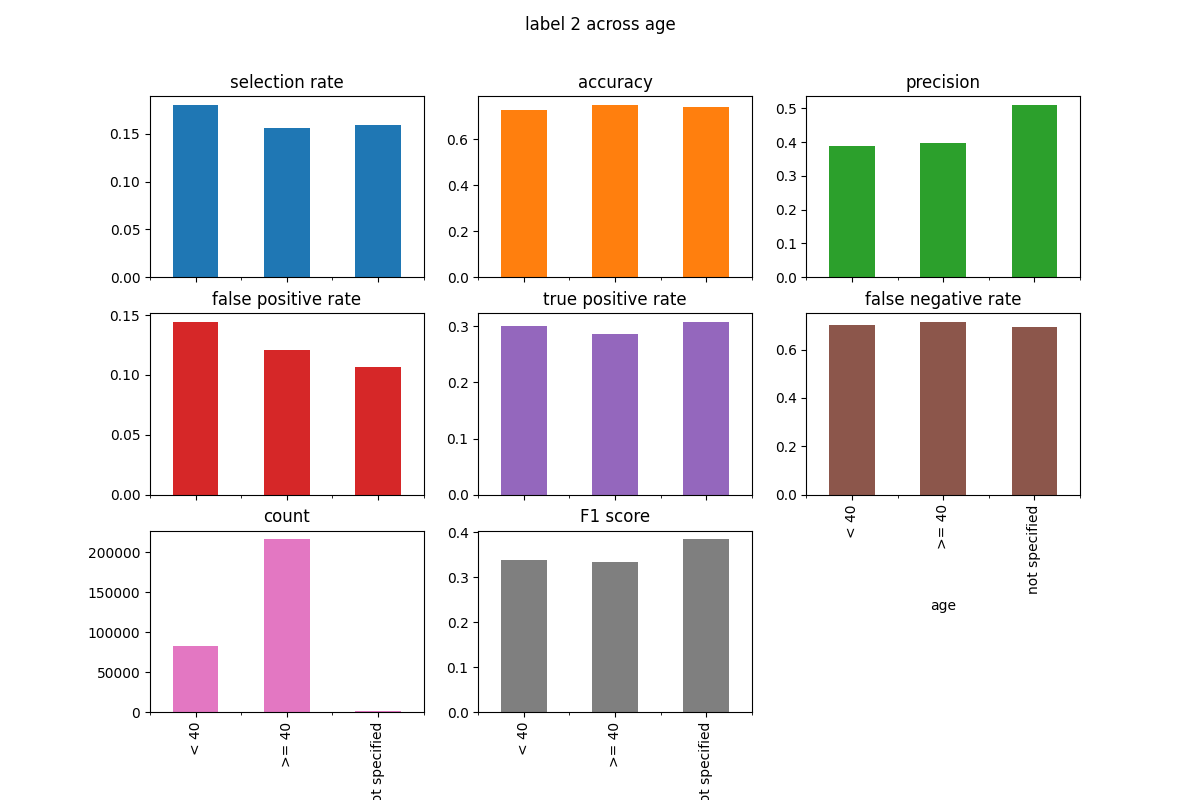}
        \caption{XGBoost classifier performance for label 2 by age group}
        \label{fig:xgb_age}
\end{figure}

\begin{figure}[ht]
    \centering 
    \includegraphics[width=1\textwidth]{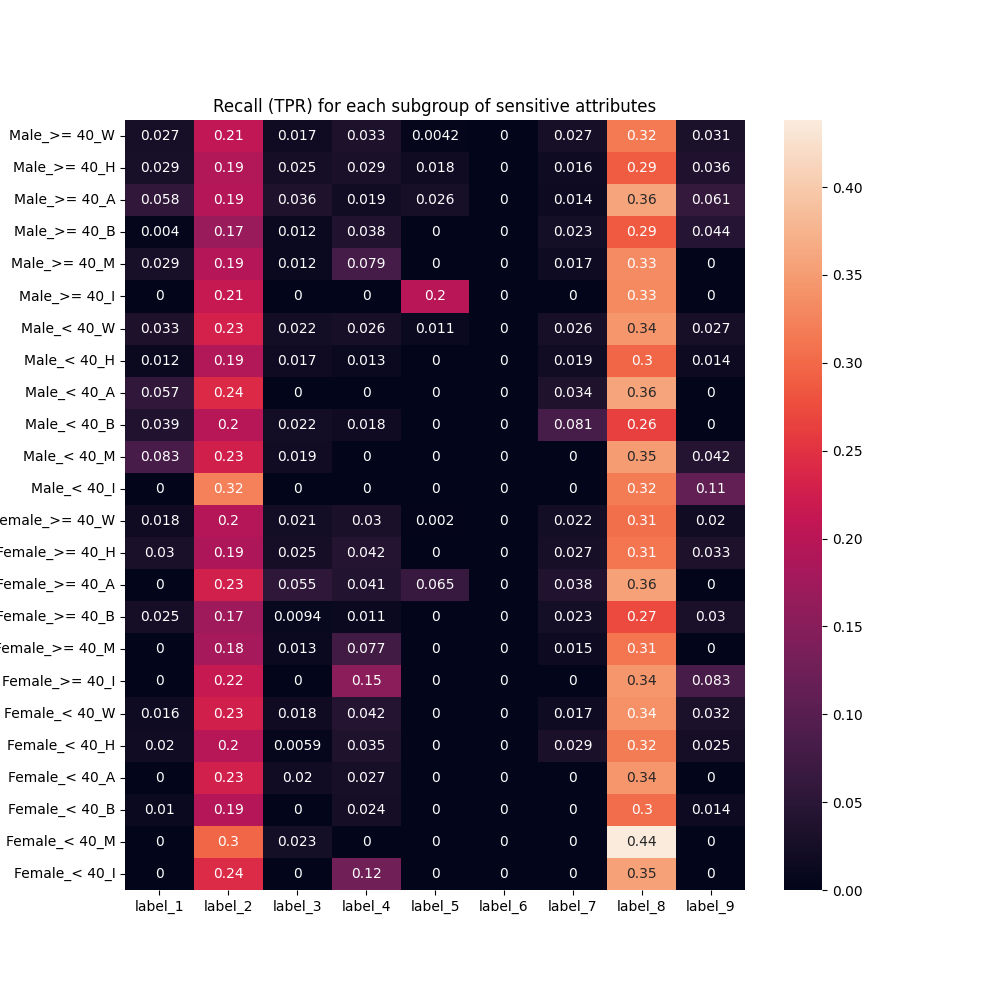}
    \caption{Recall of XGBoost for all combinations of protected groups}
    \label{fig:multilabel_xgb}
\end{figure}

To measure the model's fairness for the protected groups, \prettyref{fig:xgb_gender} and \prettyref{fig:xgb_age} show a comparison of the model's performance across gender and age groups. Additionally, we calculated the model's performance for all possible subgroups of the sensitive attributes (i.e. gender, age, and ethnicity) available in the dataset. In \prettyref{fig:multilabel_xgb}, we can see that for each label, there are visible inconsistencies in the model's performance across the subgroups. 

\begin{figure}[htbp]
\centering 
\includegraphics[width=1\textwidth]{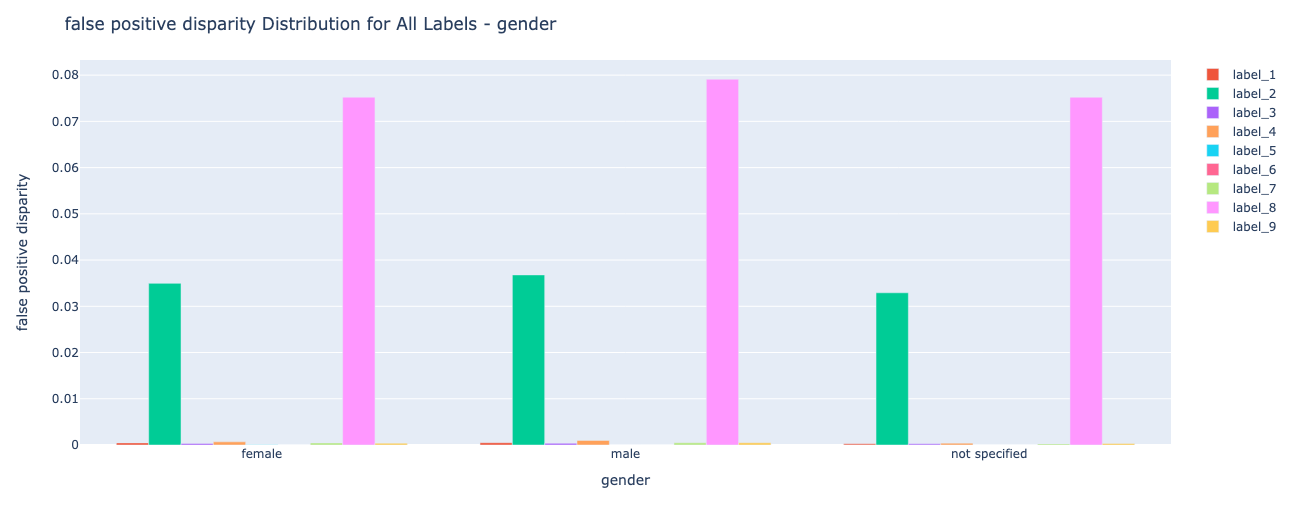}
\caption{False Positive of XGBoost across gender}
\label{multilabel_xgb}
 \end{figure}

\FloatBarrier
\FloatBarrier

\paragraph{Takeaway}

we evaluated the fairness of both a baseline logistic regression model and a more advanced XGBoost model. The logistic regression model, implemented using the OneVsRestClassifier, showed equal performance across gender groups but failed to outperform random classification. Key metrics such as accuracy, true positive rate (TPR), and false negative rate (FNR) indicated no significant disparities between male and female groups. However, the model’s overall effectiveness was limited, as it did not improve much over random predictions.

For the XGBoost classifier, the model’s performance varied across labels, with notably better recall for labels 2 and 8, likely due to a higher number of positive samples. When analyzing fairness across protected groups (gender, age, and ethnicity), we observed inconsistencies in the model’s performance. These disparities were particularly pronounced when analyzing subgroups based on sensitive attributes, as highlighted by varying performance metrics such as false positive rates and recall. This suggests that while XGBoost showed stronger overall classification performance, fairness issues persisted across subgroups.

\subsection{Multi-Label Regression}

Multi-label regression is a supervised learning problem that aims to predict multiple continuous target variables simultaneously from a common set of input features. Multi-label regression is a generalization of linear regression~\citep{borchani2015survey}.

In this section, we will apply the following strategy to measure disparities: first, we evaluate the performance of a given model without taking into account the fairness of the model. Then, the disparities in performance for each protected attribute will be evaluated. The final output is a 2-dimensional array of the performance disparities for each protected attribute.

\subsubsection{Lasso for Regression}

We first build a simple benchmark model using MultiTaskLasso. The MultiTaskLasso is a linear model that jointly estimates sparse coefficients for multiple regression problems. The constraint is that the selected features are the same for all the regression problems, also called tasks. One interesting side effect is that the model will choose the same features for all the regression problems. Therefore, we could gain some insights about which features are important for predicting all labels simultaneously.

We split the data using a stratified split with a 0.3 fraction as holdout data. We fit the model and evaluate its performance on the holdout data. We use the Mean Absolute Error as the evaluation metric. Overall, the model prediction is not very accurate, with a high MAE. This could be due to both data sparsity and the nature of imbalance labels.

However, we can see some useful insights from the model. First, the performance is much better in some label classes than in others. Second, the disparities in the performance for each protected attribute are different within each label class. This observation can help to identify or measure disparities directly. 

One extra step we could take is to perform matrix analysis on this 2D array of disparities and compare the performance disparities across different protected attributes and models. However, we do not show these results here because the measurement would be unreliable due to the lack of consistency of statistical weights of the distinct protected attributes.

\begin{figure}[htbp]
    \centering
        \centering
        \includegraphics[width=\textwidth]{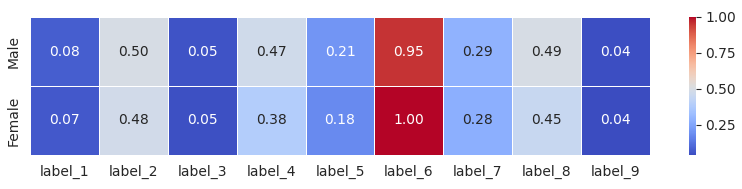}
        \caption{Disparity in Mean Absolute Error Across Gender Groups Using the Lasso Model}
        \label{fig:lasso_gender}
\end{figure}
\begin{figure}[htbp]
    \centering
        \centering
        \includegraphics[width=\textwidth]{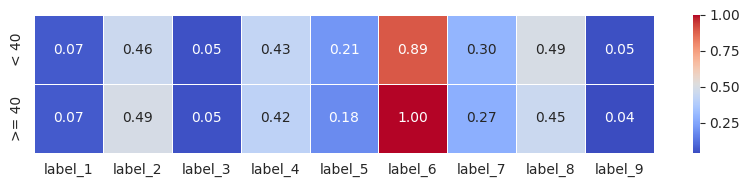}
        \caption{Disparity in Mean Absolute Error Across Age Groups Using the Lasso Model}
        \label{fig:lasso_age}
\end{figure}

    \begin{figure}[htbp]
    \centering
        \includegraphics[width=\textwidth]{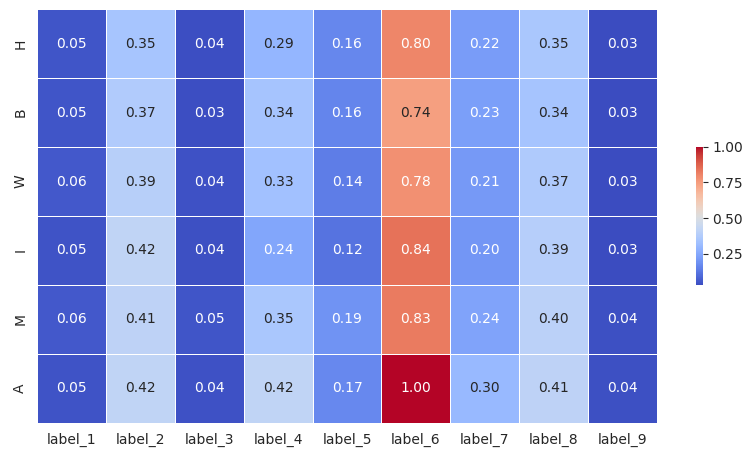}
        \caption{Disparity in Mean Absolute Error Across Ethnicity Groups Using the Lasso Model}
        \label{fig:ethn3}
    \end{figure}

\subsubsection{XGBoost for Regression}

We first trained an eXtreme Gradient Boosting (XGBoost) algorithm to measure fairness in multi-label regression. XGBoost is known for its state-of-the-art performance, computational efficiency, and built-in regularization to prevent overfitting \cite{joe2024multi}. The objective is to predict multiple continuous target variables simultaneously.

The process begins by initializing a base XGBoost regressor with default parameters and specifying the objective function as a squared error. This serves as the ensemble's foundational model. Subsequently, the base regressor is encapsulated within a MultiOutputRegressor wrapper, enabling it to handle multiple target variables.

To optimize the performance of the ensemble model, grid search cross-validation is employed. A hyperparameter grid consists of the number of estimators and the maximum depth of each estimator within the ensemble. This grid search is conducted over a 5-fold cross-validation setup, utilizing negative mean squared error (MSE) as the scoring metric to evaluate model performance.

The best estimator configuration is identified based on the specified hyperparameter grid and cross-validation results after fitting the grid search to the training data. This optimal estimator is then selected and utilized for predictions on unseen test data.

\begin{figure}[htbp]
    \centering
        \centering
        \includegraphics[width=\textwidth]{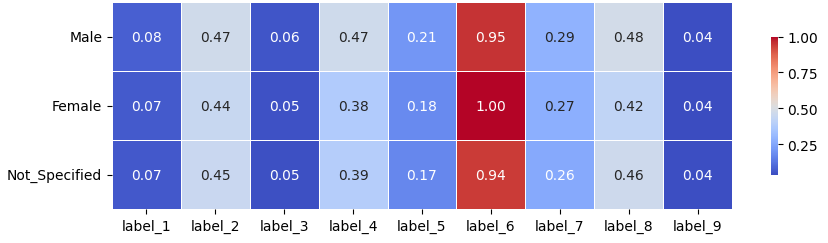}
        \caption{Disparity in Mean Absolute Error Across Gender Groups Using the XGBoost Model}
        \label{fig:gender2}
\end{figure}

\begin{figure}[htbp]
    \centering
        \centering
        \includegraphics[width=\textwidth]{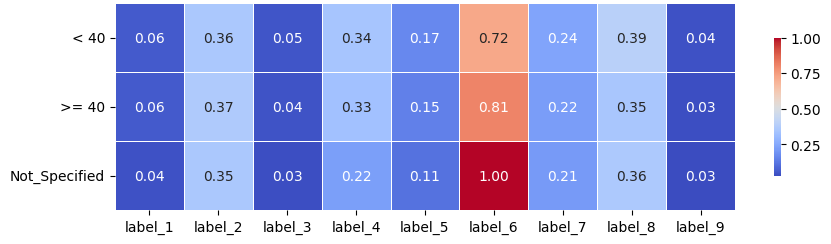}
        \caption{Disparity in Mean Absolute Error Across Age Groups Using the XGBoost Model}
        \label{fig:age2}
\end{figure}

Figure \ref{fig:gender2}  presents the mean absolute error (MAE) for a multi-label regression task using an XGBoost model, standardized across different labels and gender categories. The MAE measures how accurately the model predicts continuous outcomes, with lower values indicating better performance.

1. \textbf{Label Performance Consistency}:
   \begin{itemize}
       \item \textbf{Best Performance}: The model shows the lowest MAE values for label\_9 across all genders (Male: 0.043692, Female: 0.036748, Not\_Specified: 0.042048), indicating the highest prediction accuracy for this label.
       \item \textbf{Other Low MAE Labels}: Label\_3 also has low MAE values (Male: 0.055639, Female: 0.050878, Not\_Specified: 0.047638), suggesting good model performance.
   \end{itemize}

2. \textbf{Worst Performance}:
   \begin{itemize}
       \item \textbf{Highest MAE}: Label\_6 presents the highest MAE for all genders (Male: 0.95179, Female: 1.00000, Not\_Specified: 0.94268), indicating it is the most challenging label for the model to predict accurately.
   \end{itemize}

3. \textbf{Gender Differences}:
   \begin{itemize}
       \item \textbf{Females}: Generally, the model performs best for females, with lower MAE values across most labels compared to the other genders. Notably, the MAE for label\_6 is the highest at 1.00000.
       \item \textbf{Males}: The model has higher MAE values for males in several labels, indicating slightly poorer performance compared to females and the Not\_Specified category.

   \end{itemize}

Figure \ref{fig:age2} presents the standardized mean absolute error (MAE) for a multi-label regression task using an XGBoost model. The results are categorized by age groups: less than 40 years old (\textless 40), 40 years old and above (\textgreater= 40), and Not\_Specified. The lower the MAE, the more accurate the model's predictions.

1. \textbf{Label Performance Consistency}:
   \begin{itemize}
       \item \textbf{Best Performance}: The model shows the lowest MAE values for label\_9 across all age groups (\textless 40: 0.038114, \textgreater= 40: 0.029816, Not\_Specified: 0.025840), indicating the highest prediction accuracy for this label.
       \item \textbf{Other Low MAE Labels}: Label\_3 also has low MAE values (\textless 40: 0.046010, \textgreater= 40: 0.041085, Not\_Specified: 0.031755), suggesting good model performance.
   \end{itemize}

2. \textbf{Worst Performance}:
   \begin{itemize}
       \item \textbf{Highest MAE}: Label\_6 presents the highest MAE for all age groups (\textless 40: 0.720566, \textgreater= 40: 0.805538, Not\_Specified: 1.000000), indicating it is the most challenging label for the model to predict accurately.
   \end{itemize}

3. \textbf{Age Group Differences}:
   \begin{itemize}
       \item \textless 40: This group shows relatively low MAE values across most labels, indicating good model performance, though label\_6 remains a challenge.
       \item \textgreater= 40: The model performs comparably to the \textless 40 group, with slightly higher MAE values for some labels, but overall similar trends.
       \item Not\_Specified: This group shows the lowest MAE for several labels, particularly label\_1, label\_3, label\_4, label\_5, and label\_9, suggesting the model performs best for data where the age is not specified. However, label\_6 remains notably problematic with the highest MAE of 1.000000.
   \end{itemize}

    \begin{figure}[htbp]
    \centering
        \includegraphics[width=\textwidth]{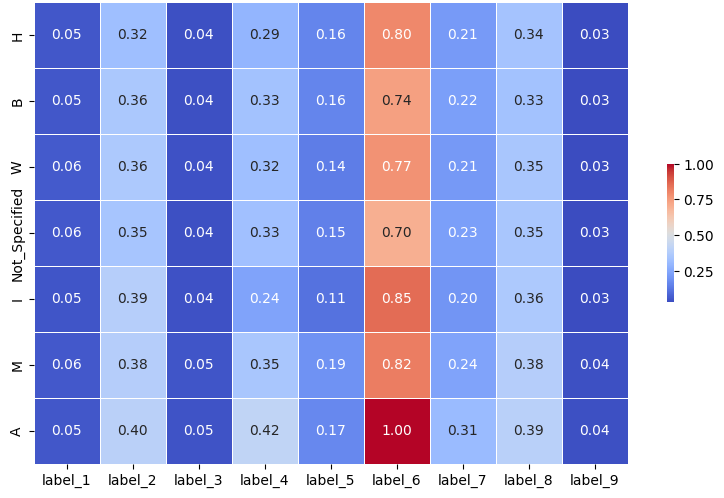}
        \caption{Disparity in Mean Absolute Error Across Ethnicity Groups Using the XGBoost Model}
        \label{fig:ethn2}
    \end{figure}

Figure \ref{fig:ethn2} presents the standardized mean absolute error (MAE) for a multi-label regression task using an XGBoost model. The results are categorized by ethnic groups, denoted as H (Hispanic), B (Black), W (White), Not\_Specified, I (Indigenous), M (Mixed), and A (Asian). The lower the MAE, the more accurate the model's predictions.

1. \textbf{Label Performance Consistency}:
   \begin{itemize}
       \item \textbf{Best Performance}: The model shows the lowest MAE values for label\_9 across most ethnic groups (H: 0.028234, B: 0.028810, W: 0.030771, Not\_Specified: 0.031807, I: 0.029145, M: 0.038112, A: 0.039556), indicating the highest prediction accuracy for this label.
       \item \textbf{Other Low MAE Labels}: Label\_3 also has low MAE values across the groups (H: 0.037374, B: 0.036100, W: 0.043217, Not\_Specified: 0.038864, I: 0.041270, M: 0.048981, A: 0.045909), suggesting good model performance.
   \end{itemize}

2. \textbf{Worst Performance}:
   \begin{itemize}
       \item \textbf{Highest MAE}: Label\_6 presents the highest MAE for all ethnic groups (H: 0.796189, B: 0.739962, W: 0.772614, Not\_Specified: 0.703195, I: 0.852938, M: 0.818958, A: 1.000000), indicating it is the most challenging label for the model to predict accurately.
   \end{itemize}

3. \textbf{Ethnic Group Differences}:
   \begin{itemize}
       \item \textbf{Hispanic (H)}: This group shows relatively low MAE values across most labels, indicating good model performance, though label\_6 remains a challenge.
       \item \textbf{Black (B)}: The model performs similarly to the Hispanic group, with slightly varying MAE values, but following the same general trend.
       \item \textbf{White (W)}: The model's performance for this group is comparable to Hispanic and Black, with slightly higher MAE values in some labels.

       \item \textbf{Indigenous (I)}: This group shows lower MAE values for labels such as label\_5 and label\_9, but label\_6 remains problematic.
       \item \textbf{Mixed (M)}: Performance for this group is generally lower than the others, especially for label\_6.
       \item \textbf{Asian (A)}: This group exhibits the highest MAE for label\_6, indicating significant challenges in prediction accuracy for this label.
   \end{itemize}

\paragraph{Takeaways}

For the multi-label regression task, we employed two models: MultiTaskLasso and XGBoost. The MultiTaskLasso model, despite its simplicity, revealed performance disparities between different protected groups. While the model’s predictive accuracy was generally low due to high mean absolute errors (MAE), it helped highlight which features were important for all regression tasks. Disparities in performance across protected groups varied by label, suggesting that some subgroups experienced higher predictive errors than others.

The XGBoost regression model, optimized using grid search, provided more accurate predictions across most labels compared to the Lasso model. However, fairness remained a challenge, as the model exhibited varying MAE across gender, age, and ethnicity subgroups. For instance, label 6 consistently showed the highest MAE across all groups, indicating difficulty in predicting this label. On the other hand, label 9 demonstrated the best performance across most protected attributes. Although XGBoost performed better than Lasso, the disparities in prediction accuracy across sensitive subgroups highlight ongoing fairness concerns in multi-label regression.

\newpage

\section{Mitigating Bias}\label{sec:mitigation}

In previous sections, we define fairness and we show how bias can be present in ML models. Several approaches have been proposed to mitigate bias~\citep{he2020geometric, dwork2012fairness, roemer2015equality,DBLP:conf/pakdd/GhodsiSN24}. In the next sections, we will explore several approaches to mitigate bias in multi-label classification and multi-label regression models.

\subsection{In-processing Mitigation}
\subsubsection{Multi-Label Classification}

\paragraph{Logistic Regression using Exponentiated Gradient Post Processing Method}
In training a fair in-processing fairness approach, we could not utilize the existing fairness methods on a multi-label classification. As such, we re-train individual classifiers for each label, and we optimize each classifier for fairness using the exponentiated gradient approach. We utilized both equalized odds and demographic parity as a constraint to the processor. In the figures below, we show the results of our respective classifier using an equalized odds and demographic parity constraint for a selected label. 
\FloatBarrier
\begin{figure}[htbp]
\centering 
\includegraphics[width=0.8\textwidth]{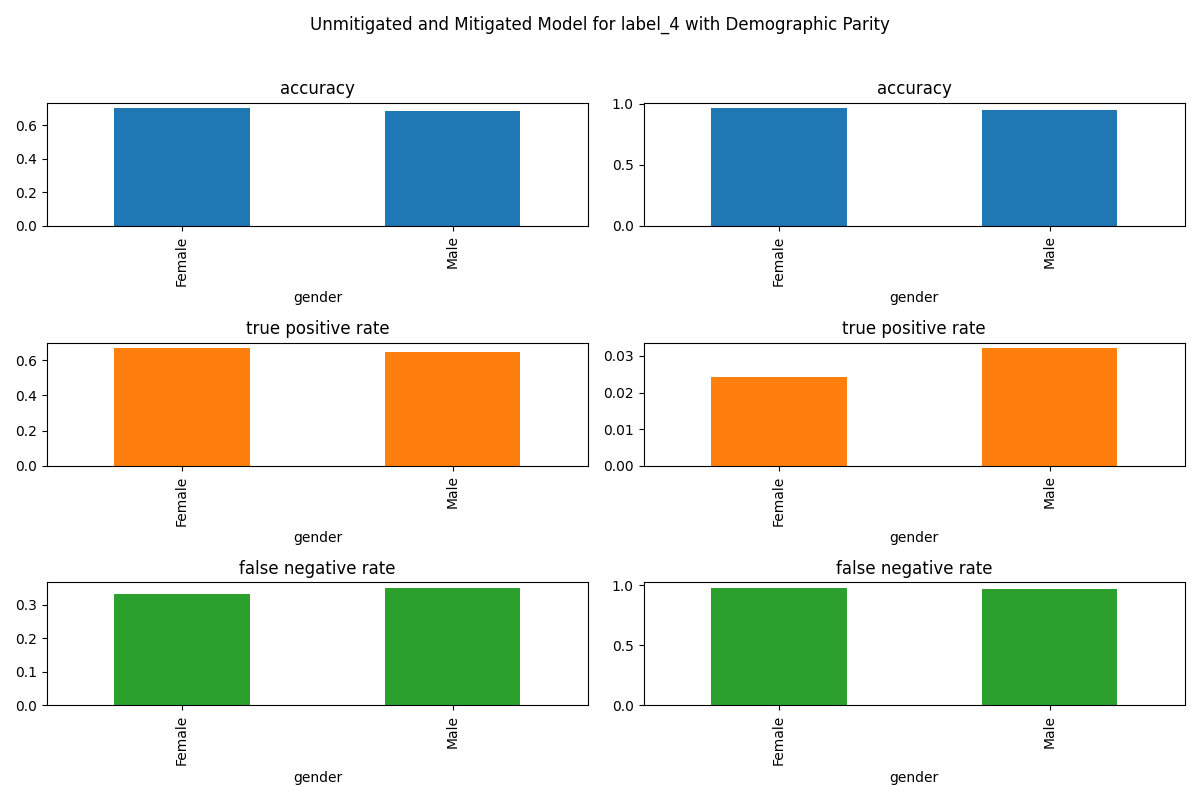}
\caption{Exponentiated gradient using Demographic parity as a constraint for label $4$}
 \end{figure}
\FloatBarrier

\FloatBarrier
 \begin{figure}[htbp]
\centering 
\includegraphics[width=0.8\textwidth]{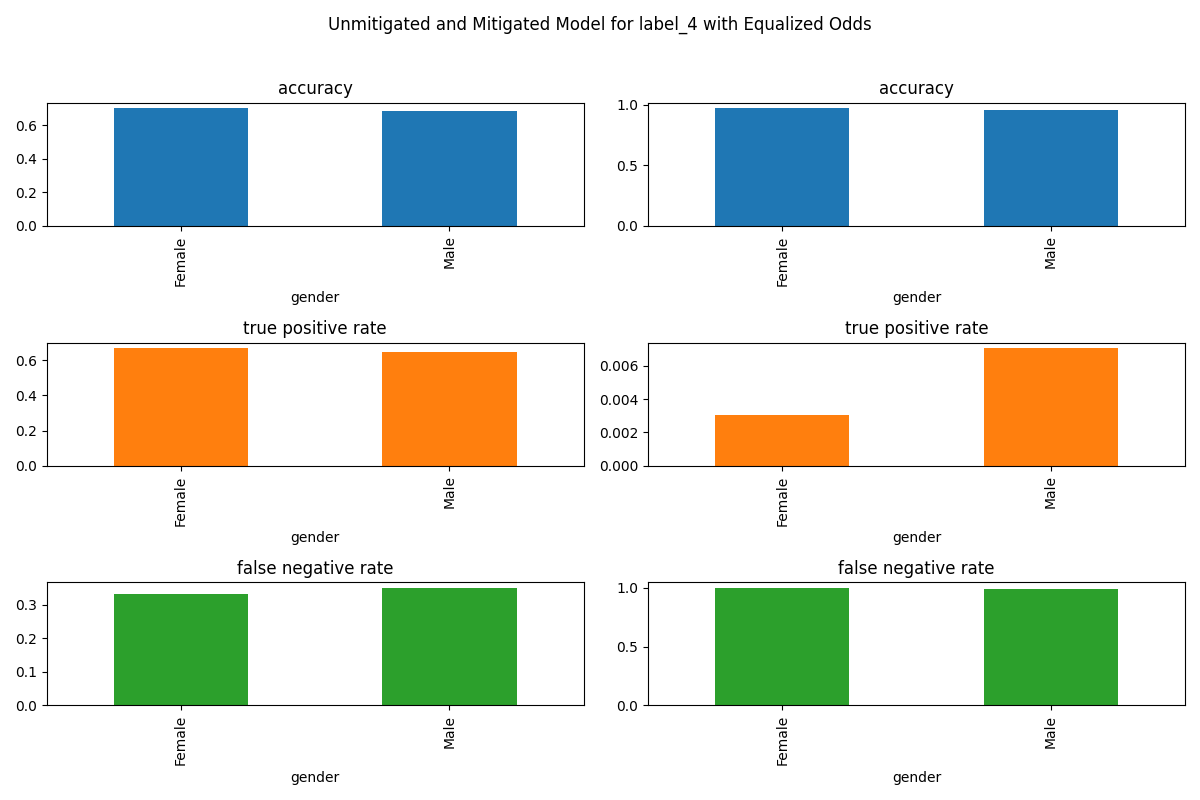}
\caption{Exponentiated gradient using Equalized odds as a constraint for label $4$}
 \end{figure}
\FloatBarrier

\subsubsection{XGBoost classifier with Exponentiated Gradient } 

Exponentiated gradient reduction(EGR)~\citep{agarwal2018reductions} is an excellent in-processing technique for reducing bias with a fairness constraint and a suitably weighted error rate as the objective. EGR aims to balance minimizing precision errors and satisfying fairness constraints.  We utilize EGR using Demographic Parity as the objective to fit the XGBoost classifier we developed.

We adopt Fairlearn~\citep{bird2020fairlearn} to mitigate the bias for classification problems.
\footnote{FairLearn is an open-source Python library developed by Microsoft, designed to facilitate the assessment and mitigation of unfairness in machine learning models. 
The goal of FairLearn is to help developers understand and improve the fairness of their machine-learning systems.}
FairLearn provides a variety of metrics to evaluate the fairness of models.  These metrics help in understanding the performance disparities across different demographic groups.

\begin{figure}[ht]
    \centering
    \subfloat[Before mitigation]{\includegraphics[width=0.8\textwidth]{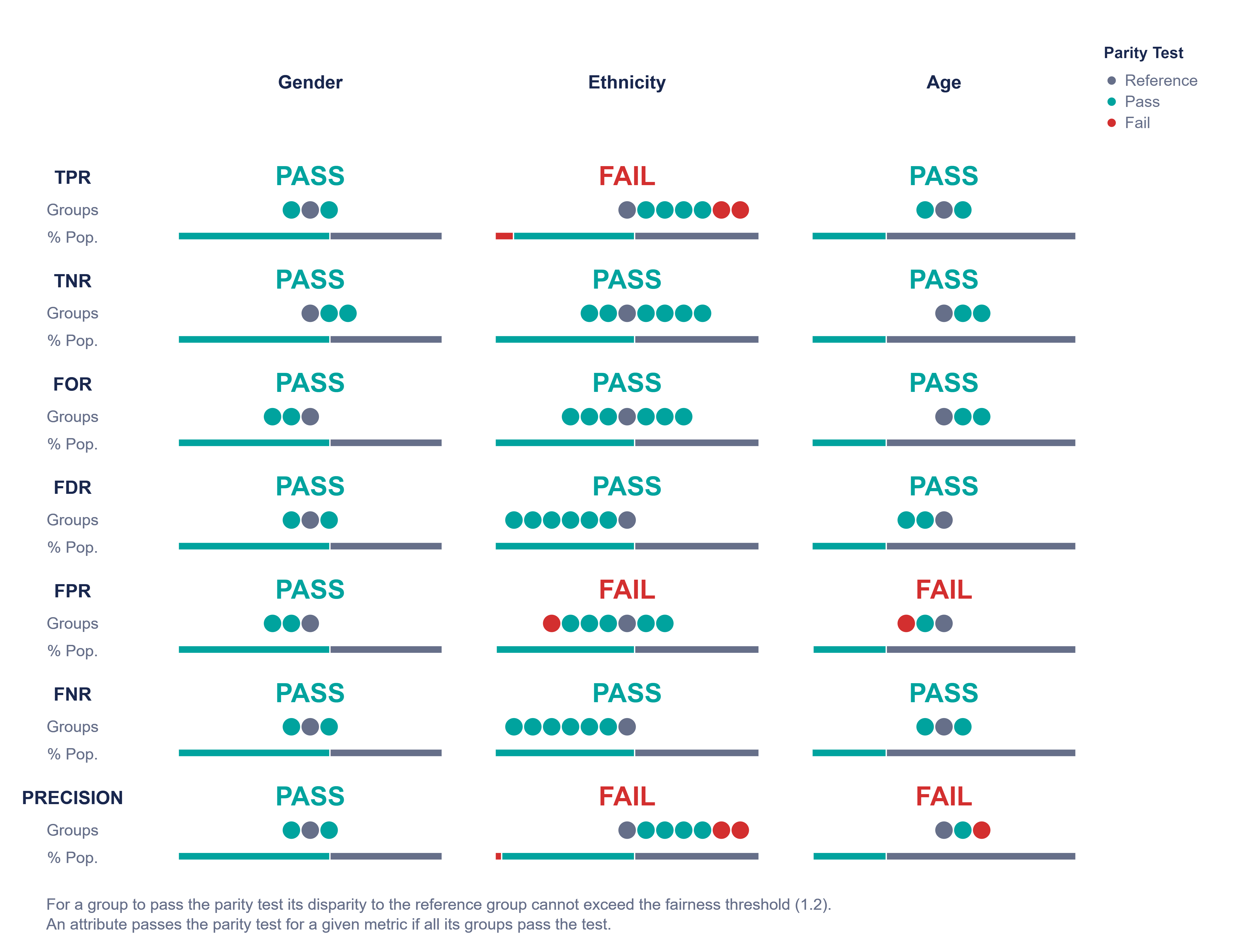}}
    \hfill
    \subfloat[After mitigation]{\includegraphics[width=0.8\textwidth]{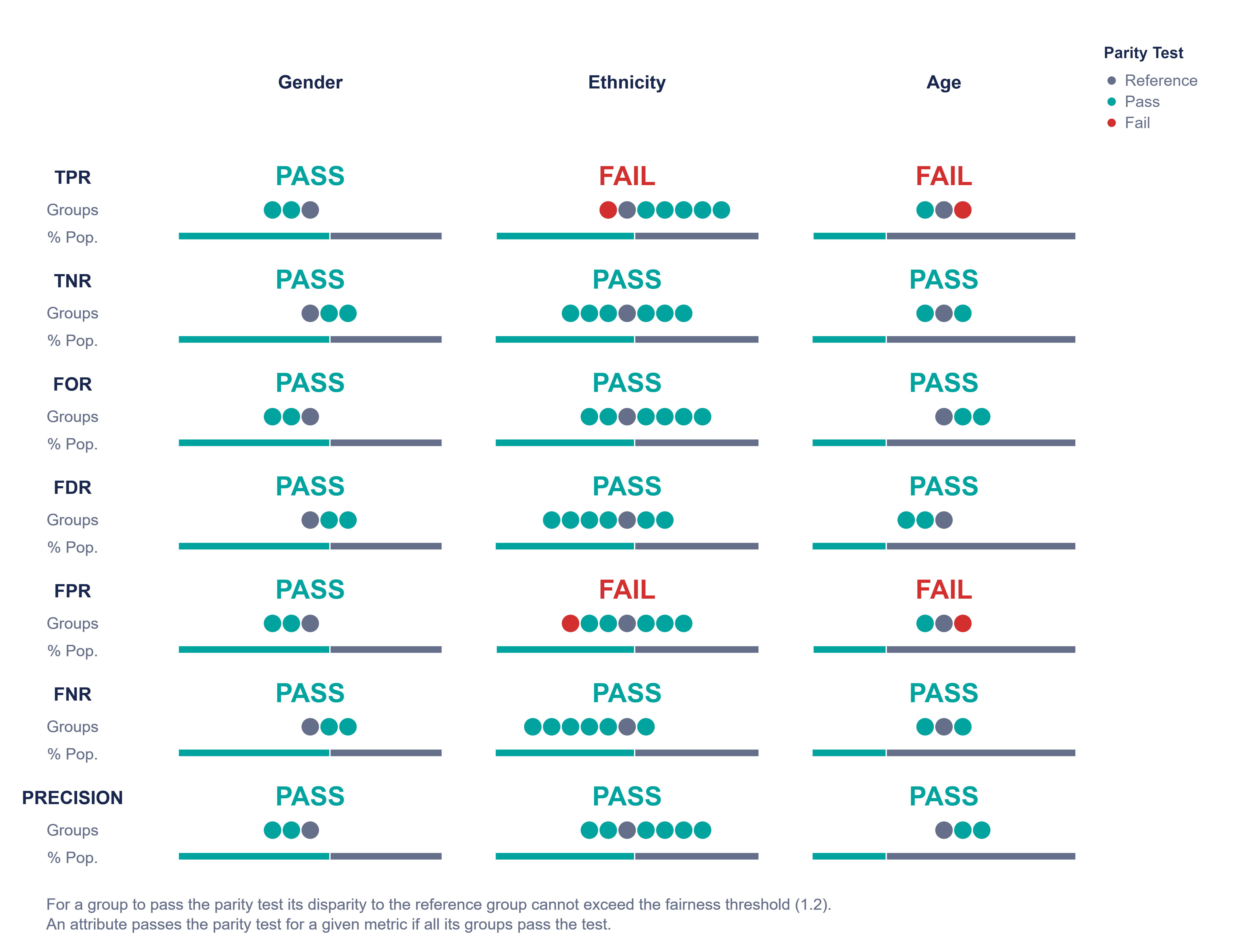}}
    \caption{}
    \label{miti_aequitas}
\end{figure}

Applying the EGR to the XGBoost classifier leads to smaller differences in disparities. Next, we utilize the Aequitas package \cite{2018aequitas} to visualize the impact of EGR on the fairness metrics in Fig. \ref{miti_aequitas}. We plot the fairness metrics across demographic groups, using 40-year-old white males as the reference group to compare the fairness measurements of minority groups. In Fig. \ref{miti_aequitas}, we display the mitigation effects of EGR on label 2 as an example. We choose a disparity threshold of 20\%. The total number of failures decreases from 5 to 4, significantly improving precision. However, there is a trade-off, as there is another failure in TPR for the age group. Overall, the positive improvements in precision and the overall failure reduction suggest progress, but the increased TPR failures highlight an area for further refinement. This is necessary to ensure the model maintains a good balance between all performance metrics while adhering to fairness standards.

\subsubsection{Using Adversarial Debiasing Method with Neural Network}
In our project, we aimed to mitigate biases in a machine learning model that offers campaigns to credit card users based on their activity over the last 12 months. Our task involved a multi-label classification problem where different labels indicated whether a cardholder had spent time in specific industries in the next three months. To ensure fairness across different sensitive groups, we implemented an adversarial learning approach inspired by \cite{DBLP:conf/aies/ZhangLM18}.

\textbf{Framework Design} Our framework consisted of two main components:

\begin{itemize}
    \item \textbf{Main Predictive Model}: This model was responsible for predicting the spending behavior of credit card users across various industries in the next three months. We experimented with a deep-learning model for this task. The model was trained as a multi-label classification model to minimize the binary cross-entropy loss between the predicted and actual labels for all industries at a time.
    \item \textbf{Adversarial Model}: The adversarial model aimed to detect and mitigate biases in the main model's predictions. It was designed to predict the sensitive attributes (such as gender or ethnicity) from the main model's output probabilities. The model was trained as a multi-label classification model to maximize the binary cross-entropy loss between the predicted and actual labels for all sensitive attributes at a time. By doing so, the adversarial model tried to ensure that the main model's predictions were not easily distinguishable by sensitive attributes, thereby promoting fairness.
\end{itemize}

\textbf{Implementation Details}

\begin{enumerate}
\item \textbf{Main Model}

We utilized a deep learning model for the main predictive task. The model architecture included multiple dense layers with ReLU activation and dropout layers for regularisation. The model was trained using binary cross-entropy loss and optimized with the Adam Optimiser.

\item \textbf{Adversarial Model}

The adversarial model was implemented using Keras. It consisted of a dense layer to predict the sensitive attribute from the main model's output and true labels. The adversarial model was designed for two fairness constraints: Demographic Parity and Equality of Opportunity.

\begin{itemize}
    \item \textbf{Demographic Parity}: The adversarial model received only the predicted labels from the main model.
    \item \textbf{Equality of Opportunity}: The adversarial model received both the predicted labels and the true labels, but was trained only on examples where the true label was the target class.
\end{itemize}

The adversarial model's architecture included:
\begin{itemize}
    \item Input layers for the predicted labels and true labels.
    \item Concatenation of these inputs followed by dense layers with sigmoid activation to predict the sensitive attribute.
    \item The adversarial loss was binary cross-entropy to maximise this loss (i.e., making it difficult to predict the sensitive attribute).
\end{itemize}
\end{enumerate}

\textbf{Combined Loss Function}

To train both models simultaneously, we defined a combined loss function:
\[
\text{Total Loss} = \text{Main Loss} + \lambda \times \text{Adversarial Loss}
\]
where $\lambda$ is a hyperparameter controlling the trade-off between the main task performance and fairness. The adversarial loss was scaled appropriately to ensure balanced training.

\textbf{Results}

We evaluated the model using standard classification metrics (accuracy, precision, recall) and fairness metrics (Demographic Parity, Equalised Opportunity, and Disparate Impact). The results showed that incorporating the adversarial model helped reduce bias, as indicated by improved fairness metrics while maintaining competitive performance on the main predictive task. We experimented with different values for the adversary loss coefficient to investigate a balancing point between fairness and model performance. 

\begin{figure}[ht]
    \centering
    \includegraphics[width=1\textwidth]{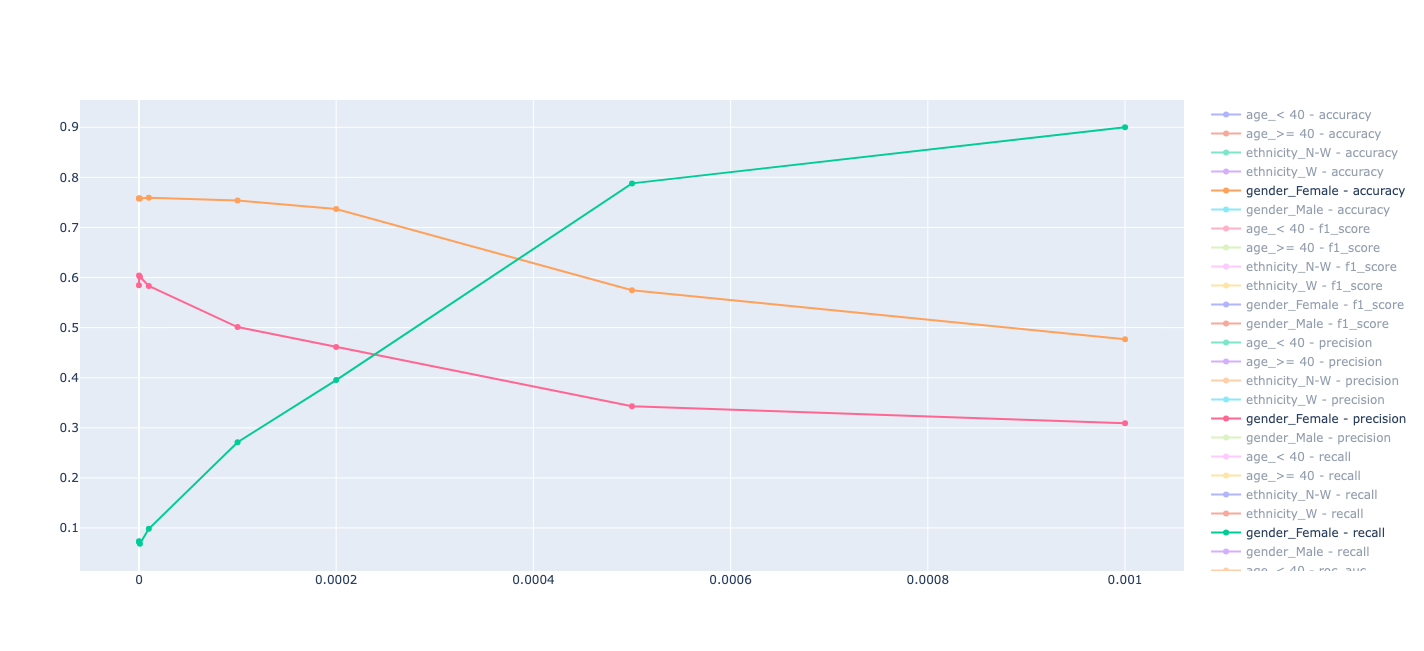}
    \caption{Accuracy, precision, and recall with varying adversarial mitigation coefficients for gender-sensitive group (Female)}
    \label{miti_adv_agg_cls_female}
\end{figure}

\begin{figure}[ht]
    \centering
    \includegraphics[width=1\textwidth]{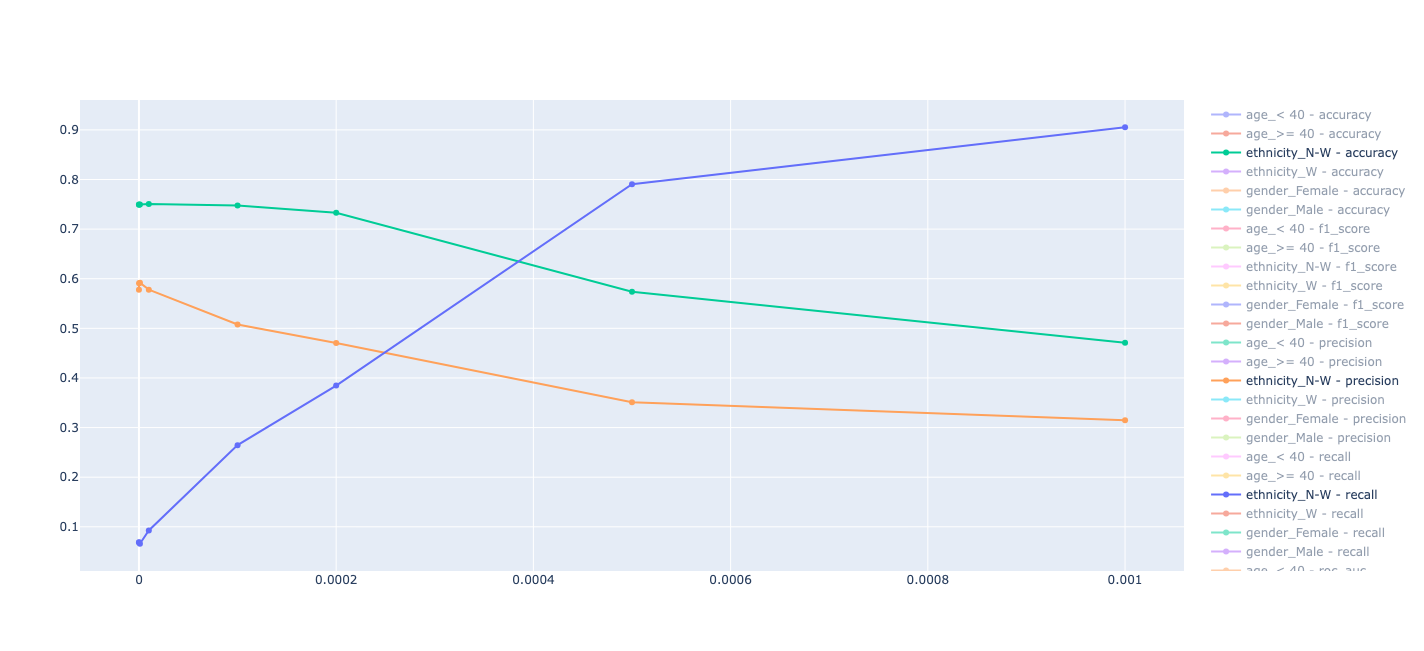}
    \caption{Accuracy, precision, and recall with varying adversarial mitigation coefficients for ethnicity-sensitive group (Non-White)}
    \label{miti_adv_agg_cls_nw}
\end{figure}

Figure \ref{miti_adv_agg_cls_female} and Figure \ref{miti_adv_agg_cls_nw} demonstrate the accuracy, precision, and recall metrics with varying adversarial mitigation coefficients for the gender and ethnicity-sensitive groups, respectively. As can be seen, the recall values increase as the mitigation coefficient increases, while the precision and accuracy values decrease for both sensitive groups. Mitigation enables the model to assign positive classes to more instances, which results in higher recall values. To find a balance in precision and recall with an acceptable accuracy, the balancing point seems to be achieved with the coefficient $0.0002$.

\begin{figure}[ht]
    \centering
    \subfloat[Before mitigation]{\includegraphics[width=1\textwidth]{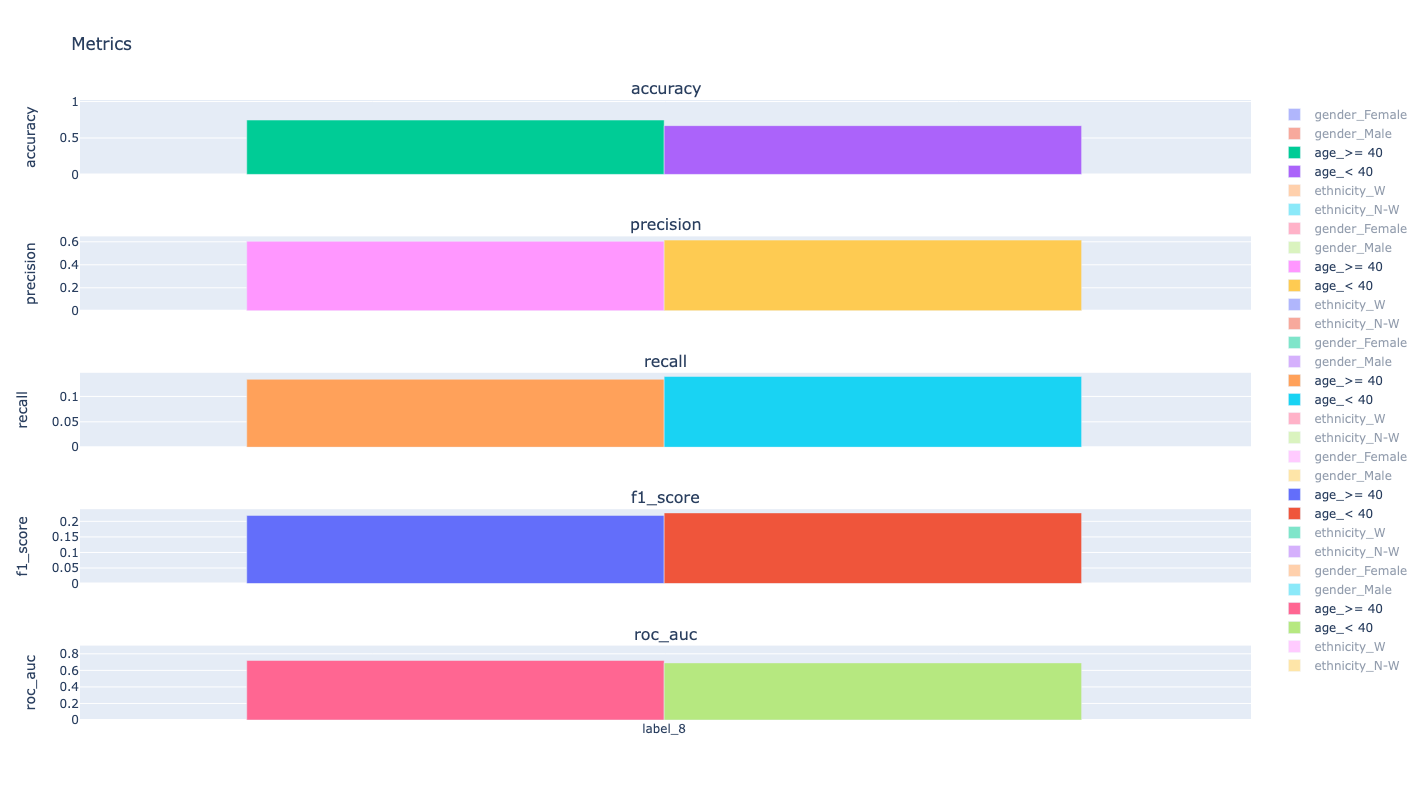}}
    \hfill
    \subfloat[After mitigation]{\includegraphics[width=1\textwidth]{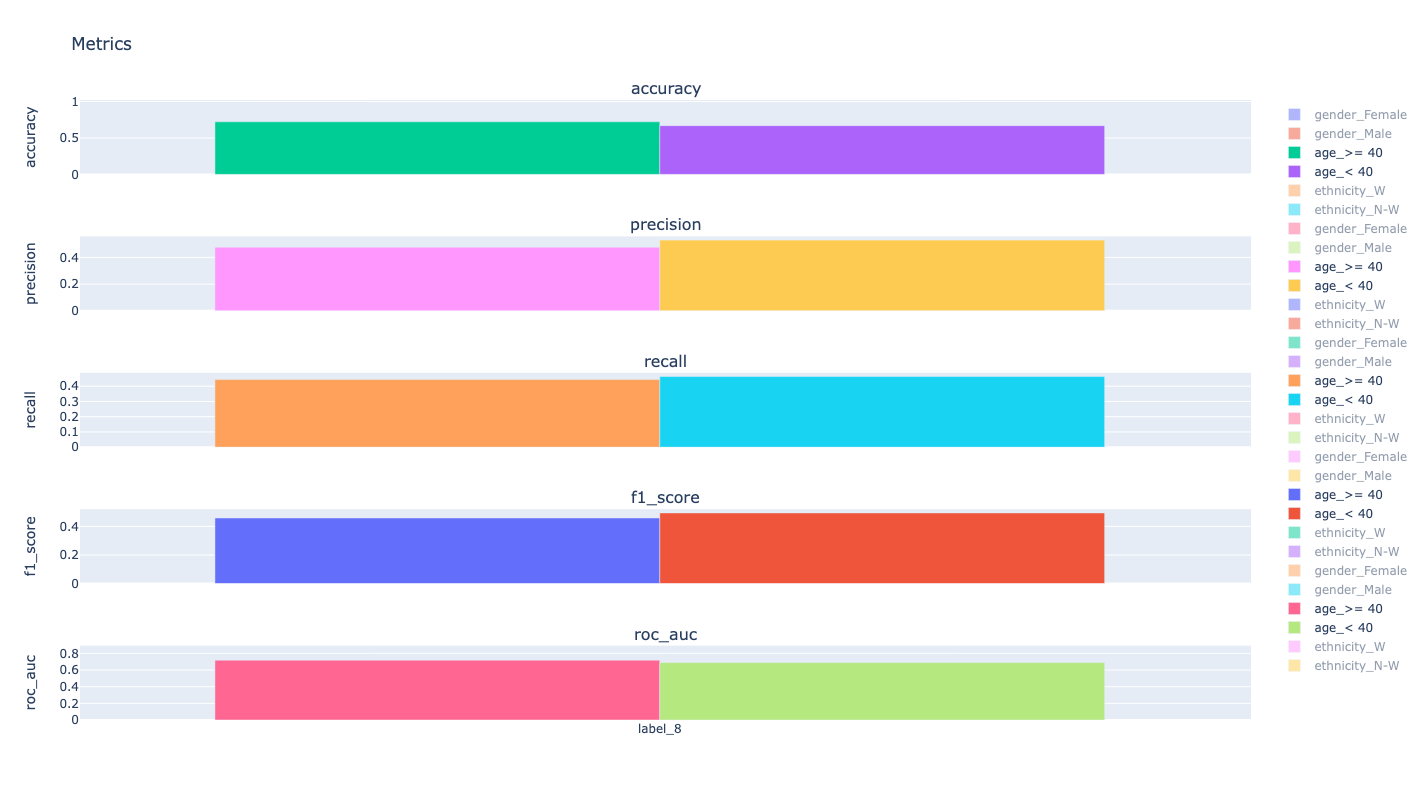}}
    \caption{Classification metrics before and after adversarial debiasing mitigation for label 8 and age groups}
    \label{miti_adv_cls_age}
\end{figure}

\clearpage
\subsubsection{Using sensitive attributes as inputs}

On the basis that sensitive attributes may contain important information for a spending adoption classifier,  we hypothesized that using the sensitive attributes as inputs could potentially improve a model's performance across protected groups and, therefore, improve performance parity (fairness).\footnote{As mentioned previously Mastercard does not include demographic features in its training data. } We are comparing fairness as unawareness, where the protected attribute is not used (current approach at Mastercard) and fairness as awareness where the protected attribute is used. By including sensitive attributes, the model could learn to control for biases represented by the sensitive attributes. Despite requiring careful implementation and consideration of ethical implications to ensure responsible and lawful use, this approach allows the model to adjust its decision thresholds or apply different strategies to ensure that the rates of positive outcomes are balanced across groups. To validate our hypothesis, we trained a multilabel XGBoost classifier that is trained on not only the 20 transactional features but also the 3 sensitive attributes. 

\paragraph{Results} 

Compared to the model trained only on the transactional features using the same hyperparameters in section \ref{multilabel_classification}, we see an improvement in the true positive rate of the model in \prettyref{tab:metric1}, particularly for labels 2 and 8. We also calculated the model's performance for all possible subgroups of the sensitive attributes (i.e., gender, age, and ethnicity) available in the dataset. In \prettyref{fig:multilabel_xgb}, we can see that for each label, there are visible improvements in the model's performance in terms of recall (true positive rate) across the subgroups.

\begin{table}[ht]
\centering
\resizebox{\textwidth}{!}{%
\begin{tabular}{cccccccc}
\toprule
label & selection rate & accuracy & precision & false positive rate & true positive rate & false negative rate  & F1 score \\
\midrule
1 & 0.016477 & 0.953177 & 0.160227 & 0.014348 & 0.074102 & 0.925898 & 0.101337 \\ 
2 & 0.183120 & 0.733113 & 0.378204 & 0.146407 & 0.311574 & 0.688426 & 0.341671 \\ 
3 & 0.020050 & 0.948337 & 0.135993 & 0.017990 & 0.073561 & 0.926439 & 0.095477 \\ 
4 & 0.015703 & 0.959130 & 0.188707 & 0.013149 & 0.095304 & 0.904696 & 0.126647 \\ 
5 & 0.002900 & 0.988227 & 0.074713 & 0.002709 & 0.023281 & 0.976719 & 0.035500 \\ 
6 & 0.000697 & 0.996160 & 0.009569 & 0.000692 & 0.002112 & 0.997888 & 0.003460 \\ 
7 & 0.011510 & 0.970407 & 0.163915 & 0.009838 & 0.086320 & 0.913680 & 0.113087 \\ 
8 & 0.265490 & 0.677937 & 0.440330 & 0.209389 & 0.402587 & 0.597413 & 0.420614 \\ 
9 & 0.009440 & 0.969920 & 0.174082 & 0.007988 & 0.068682 & 0.931318 & 0.098501 \\ 
\bottomrule
\end{tabular}
}
\caption{Performance of multilabel XGBoost classifier by labels after including sensitive attributes}
\label{tab:metric1}
\end{table}

\begin{table}[ht]
\centering
\resizebox{\textwidth}{!}{%
\begin{tabular}{cccccccc}
\toprule
label & selection rate & accuracy & precision & false positive rate & true positive rate & false negative rate & F1 score \\
\midrule
0 & -0.000477 & -0.003177 & 0.019773 & -0.001348 & 0.003898 & -0.005898 & 0.008663 \\
1 & -0.023120 & 0.006887 & 0.021796 & -0.016407 & -0.021574 & 0.021574 & -0.011671 \\
2 & -0.004050 & 0.001663 & 0.034007 & -0.003990 & 0.000439 & 0.003561 & 0.004523 \\
3 & -0.001703 & 0.000870 & 0.001293 & -0.001149 & -0.008304 & 0.005304 & -0.006647 \\
4 & -0.000200 & 0.001773 & 0.025287 & -0.000209 & 0.007719 & -0.006719 & 0.012500 \\
5 & -0.000177 & 0.003840 & 0.009431 & -0.000172 & 0.001088 & 0.002112 & 0.001940 \\
6 & -0.002810 & -0.000407 & 0.016085 & -0.002538 & -0.016320 & 0.016320 & -0.013087 \\
7 & -0.025490 & 0.012063 & 0.019670 & -0.029389 & -0.032587 & 0.032587 & -0.010614 \\
8 & 0.000260 & 0.000080 & 0.005918 & 0.000212 & 0.005318 & -0.001318 & 0.001499 \\
\bottomrule
\end{tabular}
}
\caption{Differences between not using sensitive attributes in input data (fairness as unawareness), which is the current approach at Mastercard, and including them (fairess as awareness). A negative value indicates that not using the protected attribute has improved performance. This table is the difference between~\ref{tab:metrics} (prott. att not included) and  ~\ref{tab:metric1} (prott. att. included)}
\label{tab:metris_diff}
\end{table}

\begin{figure}[htbp]
\centering 
\includegraphics[width=1\textwidth]{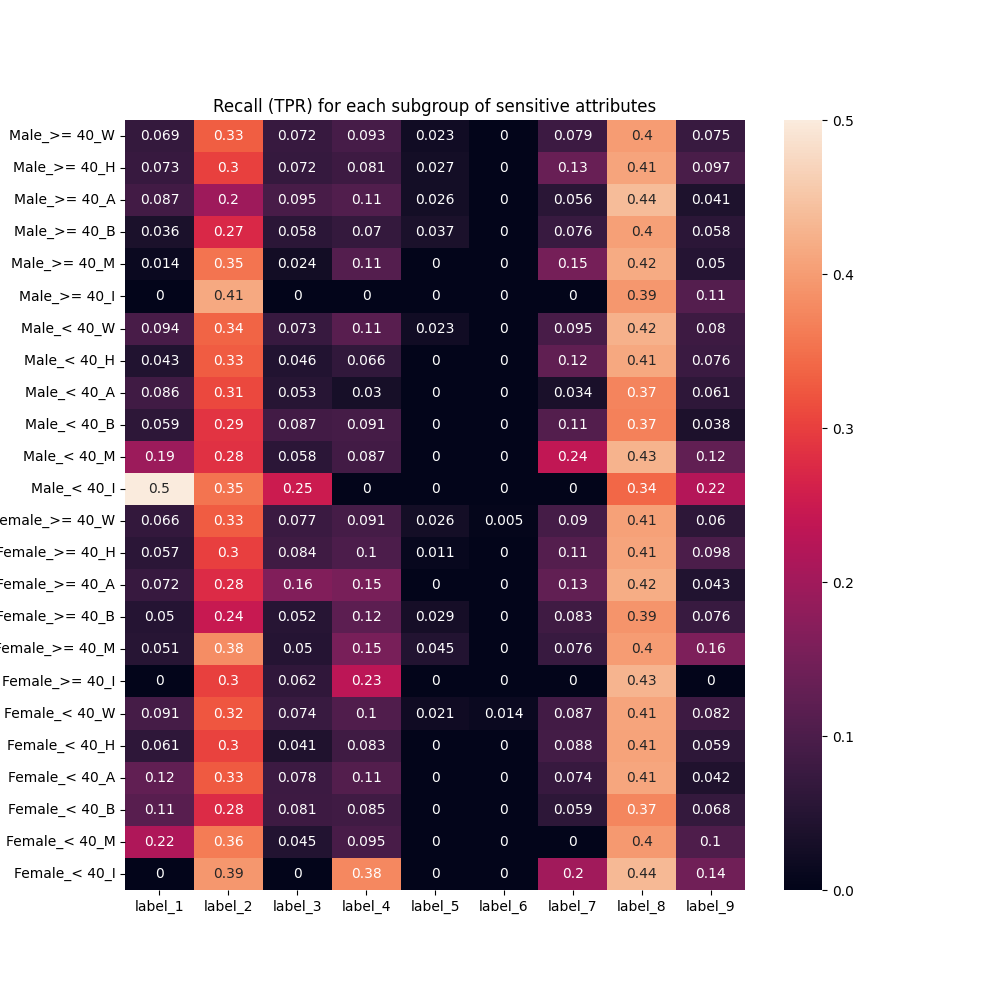}
\caption{Recall of XGBoost classifier trained on all features (including sensitive attributes) for all combinations of protected groups}
\label{fig: multilabel_xgb_sensitive}
 \end{figure}

\textbf{Preliminary Conclusions}

The results indicate that incorporating sensitive attributes into the model can improve its performance and fairness, particularly in achieving better true positive rates across various protected groups, which aligns with our initial hypothesis. As observed in Table \ref{tab:metris_diff}, the model trained with sensitive attributes shows improved recall and F1 scores for certain labels compared to the baseline model. Specifically, labels 2 and 8 demonstrated significant gains in recall, suggesting that the model can more accurately predict positive outcomes for these categories when sensitive attributes are included.

However, it is essential to acknowledge that while the inclusion of sensitive attributes appears to benefit performance and fairness, careful consideration of ethical and legal implications is crucial. The observed improvements must be weighed against the potential risks of bias perpetuation and misuse. Further analysis is needed to assess these findings across different contexts and to ensure that fairness goals are met without compromising ethical standards.

\newpage 
\subsection{Post-Processing Mitigation Assessment}

In this project, we focused on mitigating bias from three main sources: fairness-specific bias, prediction model bias, and historical bias within the data. Earlier sections of this report compiled several key metrics of fairness, which can be applied to identify the extent to which bias potentially exists in each input (i.e., features and labels). In the following sections, we aim to further explore the constructs of bias, measure their effects, and develop strategies to mitigate them.

\subsubsection{Bias Decomposition}

Our strategy involves decomposing the observed bias into components such as model bias and historical or natural bias. This decomposition is crucial because it helps us understand the different sources of bias and their impact on the predictive model. By identifying and quantifying these components, we can develop targeted interventions to mitigate bias more effectively. The process involves three main steps:

\begin{enumerate}
    \item \textbf{Use the Previously Computed Bias Metrics}:
    \begin{itemize}
        \item We use metrics such as Group-Level Demographic
Differences, Mean Difference, Disparate Impact, and Mean Squared Error (MSE) as dependent variables to quantify bias.
    \end{itemize}
    \item \textbf{Econometric Models to Determine Feature Influence}:
    \begin{itemize}
        \item Regression models aim to understand how different features influence these bias metrics. The regression equation can be represented as:
        \[
        Y = \beta_0 + \beta_1 X_1 + \beta_2 X_2 + \ldots + \beta_n X_n + \epsilon
        \]
        Here, \( Y \) represents the bias metric (e.g., Mean Difference), \( X_1, X_2, \ldots, X_n \) are the features, \( \beta_i \) are the coefficients indicating the influence of each feature, and \( \epsilon \) is the error term.
    \end{itemize}
    \item \textbf{Mitigation Strategies}:
    \begin{itemize}
        \item Based on the regression results, we apply specific mitigation techniques to reduce identified biases. By understanding the relationships between features, predicted labels, and demographic variables, we can design interventions that specifically address the sources of bias.
    \end{itemize}
\end{enumerate}

\subsubsection{Use Bias Metrics}

For instance, if our focus is on the Group-Level Demographic Differences, we compare the prediction model and the original model to measure the difference in performance metrics (e.g., accuracy, precision) across different demographic groups. This approach ensures that the model performs equitably across different segments of the population. The mathematical representation as follows:

For a demographic group \( g \):
\[
\text{Metric}_g = f(\hat{y}_g, y_g)
\]
where \( f \) represents the performance metric function (e.g., accuracy), \( \hat{y}_g \) is the predicted outcome for group \( g \), and \( y_g \) is the actual outcome for group \( g \).

The bias or disparity can be measured as:
\[
\text{Disparity} = \max_g(\text{Metric}_g) - \min_g(\text{Metric}_g)
\]
This equation calculates the range of performance across groups, highlighting the maximum disparity.

The group-level metric can also be operationalised using other metrics, depending on the context and specific fairness goals. For example: \textbf{Ratio of Positive Prediction Rates}, which measures the ratio of the rate at which positive outcomes are predicted for different groups, ensuring one group is not unfairly favoured. Additionally, \textbf{Equalised Odds Difference}, which measures the difference in true positive rates and false positive rates between groups, ensures the model performs similarly regardless of the group.

These metrics provide a more comprehensive understanding of how the model behaves across different demographics and identify areas where bias may be present.

\subsubsection{Run Regression Models to Determine Feature Influence}

Bias decomposition regression aims to understand how different features influence bias metrics. By decomposing bias into contributions from features and predicted labels, we can gain insights into the sources and magnitude of bias in a predictive model. The econometric model for bias decomposition is as follows:

\[
\text{Bias}_i = \alpha + \sum_{j=1}^{p} \beta_j X_{ij} + \sum_{k=1}^{q} \gamma_k \hat{y}_{ik} + \sum_{l=1}^{r} \delta_l D_{il} + \epsilon_i
\]

\begin{itemize}
    \item \(\text{Bias}_i\): The bias for instance \(i\). 
    \item \(X_{ij}\): Original features.
    \item \(\hat{y}_{ik}\): Predicted labels.
    \item \(D_{il}\): Demographic variables.
    \item \(\alpha\): Intercept.
    \item \(\beta_j, \gamma_k, \delta_l\): Coefficients for features, predicted labels, and demographic variables.
    \item \(\epsilon_i\): Error term.
\end{itemize}

This regression model quantifies the impact of each feature, predicted label, and demographic variable on the bias metric. By analysing the coefficients, we can determine which factors contribute most significantly to bias. Noted that, we can use previously computed bias metrics (e.g., Mean Difference, Disparate Impact, MSE) as dependent variables in the regression model. Through bias decomposition regression, we can separate the observed bias into different components:

\begin{enumerate}
    \item \textbf{Intercept \(\alpha\):}
    \begin{itemize}
        \item This represents the baseline bias when all features, predicted labels and demographic variables are zero. It's a constant that shifts the bias level up or down.
    \end{itemize}
    \item \textbf{Coefficients for Original Features \(\beta_j\):}
    \begin{itemize}
        \item Each \(\beta_j\) represents the contribution of feature \(X_j\) to the bias. A significant \(\beta_j\) indicates that feature \(X_j\) contributes to the bias. The sign (positive or negative) shows the direction of the contribution.
    \end{itemize}
    \item \textbf{Coefficients for Predicted Labels \(\gamma_k\):}
    \begin{itemize}
        \item Each \(\gamma_k\) represents the contribution of predicted label \(\hat{y}_k\) to the bias. Significant \(\gamma_k\) values indicate that the model's predictions for label \(k\) are associated with bias. This can highlight which labels are more prone to biased predictions.
    \end{itemize}
    \item \textbf{Coefficients for Demographic Variables \(\delta_l\):}
    \begin{itemize}
        \item Each \(\delta_l\) represents the contribution of the demographic variable \(D_l\) (e.g., gender, age, ethnicity) to the bias. Significant \(\delta_l\) values suggest that certain demographic groups are experiencing bias, either positive or negative.
    \end{itemize}
\end{enumerate}

By applying this analysis, we can identify the main sources of bias following the construction of our predictive models. This approach provides a comprehensive understanding of how various factors influence bias, enabling us to implement effective bias reduction techniques.

Identifying the sources of bias is crucial for effective mitigation. Significant \(\beta_j\) coefficients for certain features indicate that these features contribute to bias, suggesting the need for modification, removal, or re-weighting to reduce their impact. 

Similarly, significant \(\gamma_k\) values for predicted labels highlight which labels are problematic, guiding targeted improvements in the model for specific labels. Additionally, significant \(\delta_l\) values point to biases related to demographic groups, informing fairness interventions to ensure equitable treatment across these groups. By understanding these relationships, we can design precise strategies to mitigate bias and promote fairness in predictive models.

\subsubsection{Systematic Biases from Residuals}

Additionally, we can use residuals from the model to analyse systematic biases related to demographic features. This approach allows us to identify and quantify biases that are not captured by the main model coefficients. 

Residual for each instance:
\[
\text{Residual}_i = y_i - \hat{y}_i
\]

Regression on residuals:
\[
\text{Residual}_i = \alpha + \beta_1 \text{Gender}_i + \beta_2 \text{Age}_i + \beta_3 \text{Ethnicity}_i + \epsilon_i
\]

In this model, \(\alpha\) represents the intercept, and \(\beta_1, \beta_2, \beta_3\) are the coefficients for the demographic features. The error term \(\epsilon_i\) captures the unexplained variance. By examining these coefficients, we can determine the extent to which demographic features contribute to the residuals, indicating systematic biases.

Analysing the residuals helps in identifying patterns of bias that are not apparent from the primary model analysis. This method provides additional insights into how demographic variables influence the model's performance, enabling us to implement more effective bias mitigation strategies.

\subsubsection{Variance and Covariance Matrix}

Constructing the variance and covariance matrix for the coefficients obtained from the bias decomposition regression can provide deeper insights into the relationships between features, predicted labels, demographic variables, and their contributions to bias. This approach helps identify not only the direct contributions to bias but also the interactions and dependencies among the predictors.

The variance and covariance matrix captures the variances of individual coefficients and the covariances between pairs of coefficients. This matrix is essential for understanding how different predictors (e.g. features) interact and contribute to the overall bias in the model.

To implement this, we first obtain the coefficients \(\beta_j\), \(\gamma_k\), and \(\delta_l\) from the fitted regression model (see econometric model for bias decomposition above). Then we calculate the variance-covariance matrix of the coefficients. This matrix provides information about the variance of each coefficient and the covariance between pairs of coefficients.

Specifically, the variance of a coefficient \(\beta_j\) after estimating from the previous econometrics model, is given by:
\[
\text{Var}(\beta_j) = \sigma^2 (\mathbf{X}^T \mathbf{X})^{-1}_{jj}
\]
where \(\sigma^2\) is the variance of the residuals, and \((\mathbf{X}^T \mathbf{X})^{-1}_{jj}\) is the \(j\)-th diagonal element of the inverse of the design matrix. The covariance between two coefficients \(\beta_j\) and \(\beta_k\) is given by:
\[
\text{Cov}(\beta_j, \beta_k) = \sigma^2 (\mathbf{X}^T \mathbf{X})^{-1}_{jk}
\]
where \((\mathbf{X}^T \mathbf{X})^{-1}_{jk}\) is the \((j,k)\)-th element of the inverse of the design matrix.

For example, when we have two features, $\beta_1, \beta_2$ are the coefficients for features $X_1, X_2$. $\gamma_1$ is the coefficient for the predicted label $\hat{y}_1$. $\delta_1$ is the coefficient for the demographic variable, say gender ($D_1$). The variance-covariance matrix for these coefficients is:

\[
\Sigma = \begin{bmatrix}
\text{Var}(\beta_1) & \text{Cov}(\beta_1, \beta_2) & \text{Cov}(\beta_1, \gamma_1) & \text{Cov}(\beta_1, \delta_1) \\
\text{Cov}(\beta_1, \beta_2) & \text{Var}(\beta_2) & \text{Cov}(\beta_2, \gamma_1) & \text{Cov}(\beta_2, \delta_1) \\
\text{Cov}(\beta_1, \gamma_1) & \text{Cov}(\beta_2, \gamma_1) & \text{Var}(\gamma_1) & \text{Cov}(\gamma_1, \delta_1) \\
\text{Cov}(\beta_1, \delta_1) & \text{Cov}(\beta_2, \delta_1) & \text{Cov}(\gamma_1, \delta_1) & \text{Var}(\delta_1) \\
\end{bmatrix}
\]

The variances information can help us to assess the precision of each coefficient estimate, with higher values indicating less reliability. High covariance between features suggests multicollinearity, which can affect model stability and interpretability. Covariances between features or predicted labels and demographic variables can reveal potential biases, indicating interactions that may cause certain groups to be treated differently. 

The variance and covariance matrix is essential for identifying and mitigating sources of bias, leading to fairer and more accurate predictive models. For Mastercard, this means ensuring that customer service algorithms treat all demographic groups equitably, thereby enhancing trust and compliance with regulatory standards.

\section{Limitations and Opportunity}\label{sec:limitations}

The data exploration, modeling, experimentation, and discussions allowed for the identification of the following limitations and further research opportunities: 

\subsection{Proxy Discrimination}

Proxy discrimination presents a significant limitation in achieving fair predictive models. While sensitive attributes like demographics are intentionally excluded from the training data to uphold privacy and fairness principles, certain features and industry labels may indirectly correlate with socio-economic status. This correlation can introduce bias and lead to discrimination in the Machine Learning algorithms. 

Mastercard's current approach avoids using sensitive data directly in models (an approach known as "fairness through unawareness"). Preliminary evidence suggests that using sensitive attributes as input may improve fairness under certain conditions, but more research is needed from technical, policy, and societal perspectives.

\subsection{Fairness-Accuracy Trade-Offs}

In the context of intersectional, multi-label fairness, implementing fairness mitigation techniques can significantly impact the model's performance metrics. Balancing fairness across multiple dimensions—such as different protected attributes and labels—presents complex challenges. The first step is to assess whether improving fairness across these intersectional categories can be achieved without compromising overall model performance. One method to explore this balance is through Pareto frontiers, which help identify scenarios where both fairness and performance can be optimized simultaneously. However, achieving such an ideal balance is not always possible in multi-dimensional fairness scenarios.

If improving fairness while maintaining performance proves difficult, the next critical question becomes how much performance degradation is acceptable in order to optimize fairness. For intersectional, multi-label cases, this decision involves consideration of the trade-offs across various protected attributes and labels. In the experimental section we have seen how mitigating one fairness metric for one demographic group may increase bias (through lower model performance) towards another demographic group.

Stakeholders must collaboratively determine the acceptable performance threshold for each category in order to align with business priorities and societal values. The complexity of managing fairness in these multi-dimensional cases makes it vital to weigh performance losses against the ethical and practical benefits of fairer outcomes across the system.

\subsection{Impact of Predictions}
If a model's training data is biased, and/or a model's predictions are biased, whoever acts on those predictions may unknowingly perpetuate these biases. For instance, if a prediction is unfairly skewed towards a specific demographic, such as the white ethnic group, compared to other groups, the person acting on that prediction may inadvertently favor the white ethnic group, and thus perpetuate historical inequality. To address this issue, an end-to-end analysis across the pipeline of the predictions across different demographic groups is suggested.

\subsection{Technical challenges on multi-label and intersectional fairness}

In section \ref{sec:math.formulation} we have presented a possible formalization of the problem of fairness in multiple industries where the protected attribute has a high cardinality due to its intersectionality. We now outline two open research questions that we were not able to address as part of the Data Study Group and that could present opportunities for future research work.

\textbf{Interpretability: How to Aggregate the Fairness Matrix to a Scalar}

Interpreting a high-dimensional fairness tensor poses human interpretability challenges due to its (at least) $3-D$ representation. Aggregating the fairness matrix to a scalar would simplify the interpretation, enabling stakeholders to make informed assessments. Some potential approaches for aggregation discussed during the DSG include:

\begin{itemize}
    \item Averaging across the tensor dimensions by applying weighted sums that take into account the relative importance of each label and each protected attribute. This approach can allow for a more precise aggregation, where the contributions of individual labels and protected attributes are scaled according to Mastercard's responsible AI governance frameworks.

    \item Request Mastercard stakeholders to provide input by ranking their preferences on which industries and labels should carry more weight in the final aggregation of the fairness metric. This data can be collected to perform preference modelling, aiming that the aggregation of the fairness tensor aligns with Mastercard’s values and business objectives.

    \item Applying additional statistical measures, such as weighted averages, medians, or harmonic means, to summarize the tensor's values into a single meaningful scalar. These methods can account for the varying importance of different dimensions (e.g., labels, protected attributes, or industries).

\end{itemize}

\textbf{Evaluation: How to Compare Models or Policies Using the Tensor?}

Evaluating and comparing different models or policies using the fairness tensor requires methodologies to handle the tensor's complexity. One approach could be to compare the aggregated scalar values derived from the tensor. Alternatively, methods such as multi-dimensional scaling, tensor decomposition, optimal transport or visual analytics could be employed to compare models or policies directly in the high-dimensional space.

\section{Team members}


\textbf{\href{https://www.linkedin.com/in/carlosmougan/}{Carlos Mougan}} is a Principal Investigator at the Alan Turing Institute specialising in Model Monitoring and AI ethics. He contributed to this project by scoping it, supporting the participants and in the general write-up.

\textbf{\href{https://www.linkedin.com/in/deniz-sezin-ayvaz/}{Deniz Sezin Ayvaz}} is a data scientist with six years of academic and professional experience, specialising in critical domains such as time series forecasting, marketing effectiveness modelling, deep learning, and unsupervised learning. Her professional journey includes serving as a data science consultant to clients spanning a wide spectrum of industries, including retail, financial services, insurance, pharmaceuticals, healthcare, telecommunications, and aviation. She holds a MSc in computer engineering, and her research focus was on disease progression modelling with generative AI.

\textbf{\href{https://www.linkedin.com/in/lorenzobelenguer/}{Lorenzo Belenguer}} holds an MA in AI \& Philosophy from Northeastern University – London, and a BA in Economics and Business Sciences from the University of Valencia, Spain. Belenguer's academic pursuits primarily revolve around: ethical frameworks, bias leading to discriminatory outcomes, and human interaction with AI, to harness technology for the betterment of society at large. He published a paper suggesting a framework to detect and mitigate bias inspired by the clinical trials conducted by the pharmaceutical industry. More details here: \href{https://link.springer.com/epdf/10.1007/s43681-022-00138-8?sharing_token=HL-9sf7oVjuOdfwOQz56Eve4RwlQNchNByi7wbcMAY6536Sg1UPYTMlRh_iXkrzwpciiccDRZxm0mL5t9M43xHZu8z_zLl3kzi2vo8bekbHaumk-a57e-IMBB9r4A26AxoIoNy0e5jSLhwySMf18gwuwpWjyzgJwI0qwIEbPfyM%3D}{Link to paper}

\textbf{\href{https://www.linkedin.com/in/hurst-he/}{Hankun He}} is a Ph.D. candidate in Mathematical Sciences at Queen Mary University of London. His research interests span AI, machine learning, dynamical systems, time-series forecasting, and environmental sciences, focusing on developing advanced models to address complex real-world challenges. He has contributed significantly to several funded research projects, including those supported by the 2022 QMUL-HUST Strategic Partnership Fund, the 2023 QMUL Impact Fund "Machine Learning Algorithm for Water Quality Prediction," the 2024 QMUL ISPF-ODA Catalyse Fund "Air Pollution Statistics: Heavy Tails, Extreme Events and Health Consequences," and more, as the named data analyst in the grant applications. More details can be found on his \href{https://www.linkedin.com/in/hurst-he/}{LinkedIn profile} and in his \href{https://scholar.google.com/citations?user=_g4nbZ8AAAAJ&hl=en&oi=ao}{publications}. He contributed to this project by constructing multi-label classification models and bias mitigation models, drafting the section "Multi-Label Classification", and producing figures that visualize the impact of the mitigation methods.

\textbf{\href{https://www.linkedin.com/in/kanubalad/}{Deborah Dormah Kanubala}} is the co-organizer for the \href{https://wimlds.org/chapters/accra-ghana/}{WiMLDS Accra-Ghana} and also a PhD student at Saarland University, Germany. Her Ph.D. Research is focused on Developing Fair Machine Learning Models. Prior to this, she worked as an NLP Engineer at Proto and also lectured at Academic City University. Ms. Kanubala is a recipient of several scholarships and grants including the \href{https://mastercardfdn.org/closing-the-gender-gap-in-stem-education-in-africa/}{Mastercard Foundation Scholar grant}, \href{https://alumnode.org/alumni-projects/breaking-stem-gap-rural-africa/}{Alumnode Project grant}, and many more. She has been listed in \href{https://www.ldv.co/blog/women-spearheading-advances-in-visual-tech-and-ai?mc_cid=1a09b11732&mc_eid=82b03bca32}{120+ Women Speaheading Advances in AI}, \href{https://medium.com/women-in-ai-ethics/announcing-the-100-brilliant-women-in-ai-ethics-list-for-2024-10ed0ef02470}{100 Brilliant Women in AI Ethics list for 2024} and many more. More details here: \href{https://www.linkedin.com/in/kanubalad/}{Linkedin profile}, \href{https://kanubalad.github.io/}{personal website}. 

\textbf{\href{https://lidadadaxu.github.io}{Mingxu Li}} is a third-year PhD student at Southwest Jiaotong University, China, specialising in machine vision and visual defect detection tasks. He contributed to this project by analysing and visualising data.

\textbf{\href{https://www.linkedin.com/in/soung-low/}{Soung Low}} is a data scientist with professional experience in the financial industry in the UK. With prior experience in consumer lending, he currently focuses on the validation of machine learning models. He holds an MSc in Applied Social Data Science from the London School of Economics. 

\textbf{\href{https://faithfulco.github.io/}{Faithful Chiagoziem Onwuegbuche}} is a PhD candidate in Machine Learning at the \href{https://www.ml-labs.ie/cohort_3/faithful-onwuegbuche/}{SFI Center for Research Training in Machine Learning (ML-Labs)}, University College Dublin. His research focuses on the application of Artificial Intelligence (AI) in finance and cybersecurity. His PhD research work is on machine learning techniques for adaptive ransomware intrusion detection. He has industry experience including recently interning at Mastercard's Foundry Artificial Intelligence and Machine Learning R\&D team in Dublin, where he developed an AI model for detecting fraudulent Bitcoin transactions. Faithful holds two master’s degrees: Financial Technology (FinTech) with Distinction from the University of Stirling, UK, and Financial Mathematics from the Pan African University, Kenya, as a Commonwealth Shared Scholar and African Union Scholar, respectively. Read more about him \href{https://faithfulco.github.io/}{here}. He contributed to this work in mathematical problem formulation, exploratory data analysis, and measuring fairness in multi-label classification and regression. He also contributed to the writing of the manuscript and PowerPoint presentation slides preparation.

\textbf{\href{https://nikipi.github.io/}{Yulu Pi}} is currently a PhD researcher from University of Warwick. Her research centres on enhancing the transparency, contestability and fairness of AI by leveraging concepts and techniques from Human-Computer Interaction (HCI) and cognitive science literature. Her work extends beyond technical and design issues to consider how ethical principles can be incorporated into AI governance. She is working at The Leverhulme Centre for the Future of Intelligence at University of Cambridge on the In-Depth EU AI Act Toolkit project. 

\textbf{Natalia Sikora} is a Ph.D. candidate in Physics, with her thesis focusing on AI approaches in medical applications, early cancer detection techniques, and personalised medicine. Her work extends beyond engineering and design challenges to explore how to create fair models that are safe for both users and patients, prioritise underrepresented groups, and make diagnostics more equitable through the application of strategies that ensure robustness against data shifts or out-of-distribution scenarios. She concentrates her research on situations where underrepresented groups are systematically misdiagnosed or experience worse medical outcomes due to being under-researched, aspiring to use AI to bring us closer to tailoring treatments to the individual. Her work attempts to tackle the systemic misdiagnosis and poorer outcomes experienced by these populations. She envisions leveraging AI's transformative potential to advance truly individualised treatments, help select the best treatment approaches, and create a fairer, more effective healthcare system.

\textbf{\href{https://www.linkedin.com/in/hdan0tran/}{Dan Tran}} is an invited assistant professor at the Católica Lisbon School of Business. He holds a Ph.D. in financial economics and a M.Sc. in Risk Engineering. His research focuses on computational modelling and machine learning in banking and finance. He contributed to problem formulation, exploratory data analysis, and measuring disparities in multi-label regression.

\textbf{\href{https://vermashresth.github.io/}{Shresth Verma}} is a PhD student at Harvard University working in the domain of \href{https://teamcore.seas.harvard.edu/home}{AI for Social Impact}. His research focuses on developing fair and robust resource allocation algorithms for pressing public health challenges. Previously, he worked at Google Research India where he developed and deployed large-scale bandit algorithms to \href{https://research.google/blog/using-ml-to-boost-engagement-with-a-maternal-and-child-health-program-in-india/}{improve health literacy in underserved communities in India.} He has also worked as a data scientist in the healthcare industry with experience in modelling the health risk and wellness journey of beneficiaries. 

\textbf{\href{https://www.linkedin.com/in/hanzhi-wang-311a13171/}{Hanzhi Wang}} is a PhD student at the Cardiff University with a background in machine learning and medical imaging. He contributed to the bias mitigation methodologies.

\textbf{\href{https://www.linkedin.com/in/skyler-xie-858860252/}{Skyler Xie}} is a PhD student focusing on quantitative economics at The Alan Turing Institute and the University of Warwick. His research involves modelling dynamic changes, causal inference, and complex networks using longitudinal data, with a particular interest in modern macroeconomic datasets characterised by high-dimensional and high-frequency observations. Skyler has collaborated with public sector organisations and listed companies to develop quantitative strategies and apply innovative data science methods to address their empirical challenges.

\textbf{\href{https://www.linkedin.com/in/adeline-pelletier-phd-2b10234/}{Adeline Pelletier}} is a Data and AI Strategy lead at Mastercard. She leads methodological research work on AI fairness and its implementation inside the company and is responsible for external AI research partnerships. Prior to joining Mastercard, she was an associate professor at the University of London, teaching and conducting research on corporate strategy, financial development and digital financial services. She holds a PhD in Business Economics from the London School of Economics and a MPhil in Development Economics from the University of Cambridge.

\bibliography{references}
\bibliographystyle{apalike}

\section{Appendix}

{\small
\begin{table}[ht]
\caption{Multi-label Regression Results for Age}
\label{tab:reg_age}
\resizebox{\columnwidth}{!}{%
\begin{tabular}{|l|c|c|c|c|c|}
\hline
\textbf{index} &
  \textbf{\begin{tabular}[c]{@{}c@{}}Mean Squared \\ Error (MSE)\end{tabular}} &
  \textbf{\begin{tabular}[c]{@{}c@{}}Mean Absolute \\ Error (MAE)\end{tabular}} &
  \textbf{\begin{tabular}[c]{@{}c@{}}R-squared \\ (R2)\end{tabular}} &
  \textbf{\begin{tabular}[c]{@{}c@{}}Root Mean \\ Squared Error \\ (RMSE)\end{tabular}} &
  \textbf{\begin{tabular}[c]{@{}c@{}}Explained \\ Variance\end{tabular}} \\ \hline
(\textless{}40', 'label\_1')       & 12187.81   & 13.55  & -0.05 & 110.4   & -0.05 \\ \hline
('\&lt; 40', 'label\_2')           & 49981.59   & 80.64  & 0.02  & 223.57  & 0.03  \\ \hline
('\&lt; 40', 'label\_3')           & 4441.69    & 10.41  & -0.24 & 66.65   & -0.24 \\ \hline
('\textless 40', 'label\_4')       & 200676.01  & 77.94  & 0.01  & 447.97  & 0.01  \\ \hline
('\textless 40', 'label\_5')       & 158425.33  & 37.48  & 0.03  & 398.03  & 0.03  \\ \hline
('\textless 40', 'label\_6')       & 1392398.21 & 163.04 & -0.0  & 1180.0  & -0.0  \\ \hline
('\textless 40', 'label\_7')       & 189434.58  & 53.62  & 0.01  & 435.24  & 0.01  \\ \hline
('\textless 40', 'label\_8')       & 94822.41   & 88.65  & 0.03  & 307.93  & 0.03  \\ \hline
('\textless 40', 'label\_9')       & 2476.09    & 8.62   & -0.01 & 49.76   & -0.01 \\ \hline
('\textgreater{}= 40', 'label\_1') & 26562.96   & 13.13  & -0.01 & 162.98  & -0.01 \\ \hline
('\textgreater{}= 40', 'label\_2') & 73794.29   & 83.25  & 0.04  & 271.65  & 0.04  \\ \hline
('\textgreater{}= 40', 'label\_3') & 3419.12    & 9.3    & -0.08 & 58.47   & -0.08 \\ \hline
('\textgreater{}= 40', 'label\_4') & 238302.2   & 74.57  & 0.01  & 488.16  & 0.01  \\ \hline
('\textgreater{}= 40', 'label\_5') & 101620.75  & 33.4   & 0.01  & 318.78  & 0.01  \\ \hline
('\textgreater{}= 40', 'label\_6') & 2038991.2  & 182.27 & 0.0   & 1427.93 & 0.0   \\ \hline
('\textgreater{}= 40', 'label\_7') & 119377.24  & 49.57  & 0.01  & 345.51  & 0.01  \\ \hline
('\textgreater{}= 40', 'label\_8') & 86594.07   & 78.53  & 0.03  & 294.27  & 0.03  \\ \hline
('\textgreater{}= 40', 'label\_9') & 2477.41    & 6.75   & -0.02 & 49.77   & -0.02 \\ \hline
('NA', 'label\_1')     & 370.82     & 8.68   & -0.19 & 19.26   & -0.16 \\ \hline
('NA', 'label\_2')     & 35148.76   & 80.24  & 0.05  & 187.48  & 0.05  \\ \hline
('NA', 'label\_3')     & 834.41     & 7.19   & -0.08 & 28.89   & -0.08 \\ \hline
('NA', 'label\_4')     & 14314.92   & 50.25  & -0.03 & 119.64  & 0.01  \\ \hline
('NA', 'label\_5')     & 50969.37   & 25.61  & 0.05  & 225.76  & 0.05  \\ \hline
('NA', 'label\_6')     & 3994305.4  & 226.27 & -0.0  & 1998.58 & -0.0  \\ \hline
('NA', 'label\_7')     & 44147.3    & 47.32  & -0.01 & 210.11  & -0.01 \\ \hline
('NA', 'label\_8')     & 42237.71   & 81.0   & 0.05  & 205.52  & 0.05  \\ \hline
('NA', 'label\_9')     & 244.18     & 5.85   & -0.22 & 15.63   & -0.2  \\ \hline
\end{tabular}%
}
\end{table}

\begin{table}[ht]
\caption{Multi-label Regression Results for Gender}
\label{tab:reg_gender}
\resizebox{\columnwidth}{!}{%
\begin{tabular}{|l|c|c|c|c|c|}
\hline
\textbf{index} &
  \textbf{\begin{tabular}[c]{@{}c@{}}Mean Squared \\ Error (MSE)\end{tabular}} &
  \textbf{\begin{tabular}[c]{@{}c@{}}Mean Absolute \\ Error (MAE)\end{tabular}} &
  \textbf{\begin{tabular}[c]{@{}c@{}}R-squared \\ (R2)\end{tabular}} &
  \textbf{\begin{tabular}[c]{@{}c@{}}Root Mean \\ Squared Error \\ (RMSE)\end{tabular}} &
  \textbf{\begin{tabular}[c]{@{}c@{}}Explained \\ Variance\end{tabular}} \\ \hline
('Male', 'label\_1')           & 42621.04   & 14.47  & -0.01 & 206.45  & -0.01 \\ \hline
('Male', 'label\_2')           & 75877.48   & 85.26  & 0.03  & 275.46  & 0.03  \\ \hline
('Male', 'label\_3')           & 3354.54    & 10.09  & -0.1  & 57.92   & -0.1  \\ \hline
('Male', 'label\_4')           & 278885.44  & 84.72  & 0.01  & 528.1   & 0.01  \\ \hline
('Male', 'label\_5')           & 162277.87  & 37.47  & 0.01  & 402.84  & 0.01  \\ \hline
('Male', 'label\_6')           & 1667189.59 & 172.67 & 0.0   & 1291.2  & 0.0   \\ \hline
('Male', 'label\_7')           & 150874.95  & 52.03  & 0.01  & 388.43  & 0.01  \\ \hline
('Male', 'label\_8')           & 72159.18   & 86.83  & 0.04  & 268.62  & 0.04  \\ \hline
('Male', 'label\_9')           & 2454.22    & 7.93   & -0.01 & 49.54   & -0.01 \\ \hline
('Female', 'label\_1')         & 7822.4     & 12.3   & -0.04 & 88.44   & -0.04 \\ \hline
('Female', 'label\_2')         & 61107.62   & 80.47  & 0.04  & 247.2   & 0.04  \\ \hline
('Female', 'label\_3')         & 4057.62    & 9.23   & -0.15 & 63.7    & -0.15 \\ \hline
('Female', 'label\_4')         & 186041.77  & 68.24  & 0.0   & 431.33  & 0.0   \\ \hline
('Female', 'label\_5')         & 83908.11   & 32.22  & 0.02  & 289.67  & 0.02  \\ \hline
('Female', 'label\_6')         & 2028552.56 & 181.41 & 0.0   & 1424.27 & 0.0   \\ \hline
('Female', 'label\_7')         & 131533.14  & 49.71  & 0.01  & 362.67  & 0.01  \\ \hline
('Female', 'label\_8')         & 93922.14   & 76.56  & 0.02  & 306.47  & 0.02  \\ \hline
('Female', 'label\_9')         & 2503.85    & 6.67   & -0.01 & 50.04   & -0.01 \\ \hline
('NA', 'label\_1') & 6592.4     & 12.12  & -0.17 & 81.19   & -0.17 \\ \hline
('NA', 'label\_2') & 61097.85   & 81.5   & 0.04  & 247.18  & 0.04  \\ \hline
('NA', 'label\_3') & 1567.87    & 8.64   & -0.29 & 39.6    & -0.29 \\ \hline
('NA', 'label\_4') & 250323.42  & 70.58  & 0.01  & 500.32  & 0.01  \\ \hline
('NA', 'label\_5') & 45255.3    & 31.23  & 0.02  & 212.73  & 0.02  \\ \hline
('NA', 'label\_6') & 2067367.89 & 171.01 & 0.0   & 1437.83 & 0.0   \\ \hline
('NA', 'label\_7') & 55959.5    & 47.16  & -0.0  & 236.56  & -0.0  \\ \hline
('NA', 'label\_8') & 227946.4   & 83.89  & 0.01  & 477.44  & 0.01  \\ \hline
('NA', 'label\_9') & 2128.14    & 7.63   & -0.29 & 46.13   & -0.29 \\ \hline
\end{tabular}%
}
\end{table}


\begin{longtable}{|l|c|c|c|c|c|}
\caption{Multi-label Regression Results for Ethnicity}
\label{tab:reg_ethnicity}\\
\hline
\textbf{index} &
  \textbf{\begin{tabular}[c]{@{}c@{}}Mean Squared \\ Error (MSE)\end{tabular}} &
  \textbf{\begin{tabular}[c]{@{}c@{}}Mean Absolute \\ Error (MAE)\end{tabular}} &
  \textbf{\begin{tabular}[c]{@{}c@{}}R-squared \\ (R2)\end{tabular}} &
  \textbf{\begin{tabular}[c]{@{}c@{}}Root Mean \\ Squared Error \\ (RMSE)\end{tabular}} &
  \textbf{\begin{tabular}[c]{@{}c@{}}Explained \\ Variance\end{tabular}} \\ \hline
\endfirsthead
\multicolumn{6}{c}%
{{\bfseries Table \thetable\ continued from previous page}} \\
\hline
\textbf{index} &
  \textbf{\begin{tabular}[c]{@{}c@{}}Mean Squared \\ Error (MSE)\end{tabular}} &
  \textbf{\begin{tabular}[c]{@{}c@{}}Mean Absolute \\ Error (MAE)\end{tabular}} &
  \textbf{\begin{tabular}[c]{@{}c@{}}R-squared \\ (R2)\end{tabular}} &
  \textbf{\begin{tabular}[c]{@{}c@{}}Root Mean \\ Squared Error \\ (RMSE)\end{tabular}} &
  \textbf{\begin{tabular}[c]{@{}c@{}}Explained \\ Variance\end{tabular}} \\ \hline
\endhead
('H', 'label\_1')              & 5926.5     & 12.52  & -0.14 & 76.98   & -0.14 \\ \hline
('H', 'label\_2')              & 40514.09   & 74.77  & 0.05  & 201.28  & 0.05  \\ \hline
('H', 'label\_3')              & 2843.47    & 8.72   & -0.16 & 53.32   & -0.16 \\ \hline
('H', 'label\_4')              & 165759.7   & 67.22  & 0.01  & 407.14  & 0.01  \\ \hline
('H', 'label\_5')              & 144666.13  & 37.52  & 0.05  & 380.35  & 0.05  \\ \hline
('H', 'label\_6')              & 1911338.2  & 185.69 & 0.0   & 1382.51 & 0.0   \\ \hline
('H', 'label\_7')              & 144140.81  & 49.7   & 0.0   & 379.66  & 0.0   \\ \hline
('H', 'label\_8')              & 61566.99   & 78.37  & 0.03  & 248.13  & 0.03  \\ \hline
('H', 'label\_9')              & 2285.43    & 6.58   & -0.01 & 47.81   & -0.01 \\ \hline
('B', 'label\_1')              & 2939.12    & 11.93  & -0.26 & 54.21   & -0.26 \\ \hline
('B', 'label\_2')              & 109714.31  & 83.04  & 0.02  & 331.23  & 0.02  \\ \hline
('B', 'label\_3')              & 1907.17    & 8.42   & -0.19 & 43.67   & -0.19 \\ \hline
('B', 'label\_4')              & 273111.86  & 77.98  & 0.0   & 522.6   & 0.0   \\ \hline
('B', 'label\_5')              & 242732.32  & 36.69  & 0.0   & 492.68  & 0.0   \\ \hline
('B', 'label\_6')              & 1774032.59 & 172.58 & -0.0  & 1331.93 & -0.0  \\ \hline
('B', 'label\_7')              & 164572.86  & 51.74  & 0.01  & 405.68  & 0.01  \\ \hline
('B', 'label\_8')              & 79119.64   & 77.51  & 0.02  & 281.28  & 0.02  \\ \hline
('B', 'label\_9')              & 1293.97    & 6.72   & -0.05 & 35.97   & -0.05 \\ \hline
('W', 'label\_1')              & 10848.07   & 13.76  & -0.03 & 104.15  & -0.03 \\ \hline
('W', 'label\_2')              & 77304.94   & 84.48  & 0.04  & 278.04  & 0.04  \\ \hline
('W', 'label\_3')              & 4352.16    & 10.08  & -0.15 & 65.97   & -0.15 \\ \hline
('W', 'label\_4')              & 236170.91  & 75.3   & 0.01  & 485.97  & 0.01  \\ \hline
('W', 'label\_5')              & 81603.18   & 32.87  & 0.01  & 285.66  & 0.01  \\ \hline
('W', 'label\_6')              & 1983222.53 & 180.19 & 0.0   & 1408.27 & 0.0   \\ \hline
('W', 'label\_7')              & 106761.43  & 48.37  & 0.01  & 326.74  & 0.01  \\ \hline
('W', 'label\_8')              & 88884.95   & 81.12  & 0.03  & 298.14  & 0.03  \\ \hline
('W', 'label\_9')              & 1819.8     & 7.18   & -0.03 & 42.66   & -0.03 \\ \hline
('NA', 'label\_1') & 57432.93   & 12.96  & -0.01 & 239.65  & -0.01 \\ \hline
('NA', 'label\_2') & 49492.72   & 80.48  & 0.04  & 222.47  & 0.04  \\ \hline
('NA', 'label\_3') & 2670.15    & 9.06   & -0.08 & 51.67   & -0.08 \\ \hline
('NA', 'label\_4') & 228191.59  & 76.01  & 0.01  & 477.69  & 0.01  \\ \hline
('NA', 'label\_5') & 133840.98  & 34.54  & 0.0   & 365.84  & 0.0   \\ \hline
('NA', 'label\_6') & 1517945.79 & 164.0  & 0.0   & 1232.05 & 0.0   \\ \hline
('NA', 'label\_7') & 183713.43  & 52.61  & 0.01  & 428.62  & 0.01  \\ \hline
('NA', 'label\_8') & 100950.51  & 81.68  & 0.03  & 317.73  & 0.03  \\ \hline
('NA', 'label\_9') & 2799.29    & 7.42   & -0.02 & 52.91   & -0.02 \\ \hline
('I', 'label\_1')              & 1561.02    & 11.81  & -0.3  & 39.51   & -0.29 \\ \hline
('I', 'label\_2')              & 82014.45   & 90.3   & 0.05  & 286.38  & 0.05  \\ \hline
('I', 'label\_3')              & 1680.56    & 9.63   & -0.86 & 40.99   & -0.86 \\ \hline
('I', 'label\_4')              & 24984.86   & 55.06  & 0.03  & 158.07  & 0.04  \\ \hline
('I', 'label\_5')              & 33363.43   & 26.15  & 0.0   & 182.66  & 0.0   \\ \hline
('I', 'label\_6')              & 2093255.72 & 198.92 & 0.0   & 1446.81 & 0.0   \\ \hline
('I', 'label\_7')              & 83325.26   & 45.96  & 0.01  & 288.66  & 0.01  \\ \hline
('I', 'label\_8')              & 105746.04  & 83.93  & 0.04  & 325.19  & 0.04  \\ \hline
('I', 'label\_9')              & 896.09     & 6.8    & 0.02  & 29.93   & 0.02  \\ \hline
('M', 'label\_1')              & 12712.89   & 13.42  & -0.01 & 112.75  & -0.01 \\ \hline
('M', 'label\_2')              & 77854.88   & 89.58  & 0.05  & 279.02  & 0.05  \\ \hline
('M', 'label\_3')              & 10563.47   & 11.42  & -0.01 & 102.78  & -0.01 \\ \hline
('M', 'label\_4')              & 178619.9   & 81.63  & 0.02  & 422.63  & 0.02  \\ \hline
('M', 'label\_5')              & 245932.94  & 44.21  & 0.03  & 495.92  & 0.03  \\ \hline
('M', 'label\_6')              & 1890167.95 & 191.0  & 0.01  & 1374.83 & 0.01  \\ \hline
('M', 'label\_7')              & 77247.89   & 55.08  & 0.01  & 277.94  & 0.01  \\ \hline
('M', 'label\_8')              & 89689.14   & 89.14  & 0.04  & 299.48  & 0.04  \\ \hline
('M', 'label\_9')              & 8294.06    & 8.89   & 0.01  & 91.07   & 0.01  \\ \hline
('A', 'label\_1')              & 2747.73    & 12.17  & -0.61 & 52.42   & -0.6  \\ \hline
('A', 'label\_2')              & 68567.48   & 92.64  & 0.04  & 261.85  & 0.04  \\ \hline
('A', 'label\_3')              & 4436.16    & 10.71  & -0.2  & 66.6    & -0.2  \\ \hline
('A', 'label\_4')              & 282795.72  & 96.98  & 0.01  & 531.79  & 0.01  \\ \hline
('A', 'label\_5')              & 73491.1    & 39.23  & 0.01  & 271.09  & 0.01  \\ \hline
('A', 'label\_6')              & 3804276.89 & 233.22 & 0.0   & 1950.46 & 0.0   \\ \hline
('A', 'label\_7')              & 193752.24  & 71.85  & 0.01  & 440.17  & 0.01  \\ \hline
('A', 'label\_8')              & 81142.71   & 91.85  & 0.04  & 284.86  & 0.04  \\ \hline
('A', 'label\_9')              & 11130.6    & 9.23   & 0.0   & 105.5   & 0.0   \\ \hline
\end{longtable}
}

\end{document}